\documentclass[preprint,12pt]{elsarticle}

\usepackage{amsmath,amssymb}
\usepackage{graphicx}
\usepackage{booktabs}
\usepackage{xcolor}
\usepackage{float}
\usepackage{algorithm}
\usepackage{algpseudocode}
\usepackage{rotating}
\usepackage{pdflscape}
\biboptions{sort&compress}

\begin{document}

\begin{frontmatter}

\title{Mask-Morph Graph U-Net: A Generalisable Mesh-Based Surrogate for Crashworthiness Field Prediction under Large Geometric Variation}



\author[label1]{Haoran Li} 
\author[label2]{Tobias Lehrer}
\author[label1]{Yingxue Zhao}
\author[label1]{Haosu Zhou}
\author[label3]{Philipp Stocker}
\author[label4]{Tobias Pfaff}
\author[label3]{Marcus Wagner}
\author[label1]{Nan Li\corref{cor}}

\cortext[cor]{Corresponding author}
\ead{n.li09@imperial.ac.uk}

\affiliation[label1]{organization={Dyson School of Design Engineering, Imperial College London},
            city={London},
            country={UK}}

\affiliation[label2]{organization={TUM School of Engineering and Design, Technical University of Munich},
            city={Munich},
            country={Germany}}

\affiliation[label3]{organization={Faculty of Mechanical Engineering, OTH Regensburg},
            city={Regensburg},
            country={Germany}}

\affiliation[label4]{organization={NVIDIA},
            country={UK}}


\begin{abstract}
Nonlinear finite element crash simulations are accurate but computationally expensive, limiting their use in iterative design optimisation. Machine-learning surrogate models based on graph neural networks (GNNs) offer a faster alternative. Message-passing GNNs are widely used for mesh simulation, and their shared node and edge update functions are relatively generalisable across varying graph structures. By contrast, non-shareable edge-specific aggregation layers can capture nonlinear relationships more accurately but usually require fixed graph connectivity, which limits generalisability. This paper presents Mask-Morph Graph U-Net (MMGUNet), a practical approach to addressing the limitation of hierarchical Graph U-Net architectures that use edge-specific downsampling and upsampling layers. Fixed coarse graph connectivity is required for edge-specific layers. To retain this while improving spatial correspondence, the proposed method morphs the coarsened graph hierarchy to each input mesh using feature-aligned barycentric parameterisation before constructing cross-graph edges. It further applies node masking during supervised pretraining, followed by parameter-efficient fine-tuning in which high-parameter edge-specific layers are frozen. The proposed approach is evaluated in in-distribution, out-of-distribution, and cross-component transfer settings using mean Euclidean distance and maximum intrusion percentage error. Results show that coarse-graph morphing improves test accuracy relative to a fixed-coarse-graph baseline, while masked supervised pretraining reduces the train-test discrepancy and improves data efficiency during transfer. The proposed model also achieves lower prediction error than external baselines. These results demonstrate a practical route toward reusable, data-efficient mesh-based surrogate modelling for crashworthiness design exploration.\end{abstract}


\end{frontmatter}

\section{Introduction}
\label{Introduction}

Crashworthiness is an important performance criterion in the structural design of safety-critical vehicle components, as it measures their ability to protect passengers during vehicle accidents. Crashworthiness analysis is traditionally performed via nonlinear finite element (FE) simulations that capture complex crash modes with large deformations \cite{Wu2006CrashFE,Chang2007CrashFE}. Despite their accuracy, such FE analyses are computationally expensive, which limits their application in iterative workflows of design optimisation. This motivates the development of machine-learning surrogate models. Early surrogate models for crashworthiness largely focused on predicting scalar responses like peak crushing force and specific energy absorption \cite{Albak2023MultiCell,Xiong2018SideStructure,AhmadiDastjerdi2019BiThin,Zende2022CompositeTube}, using simple architectures such as multilayer perceptrons (MLPs) and recurrent neural networks (RNNs) \cite{Rogala2020Nonlinearity,Sakaridis2022TubeML,Kohar2020Lightweighting,Guo2023SubwayCone}. These models are limited to scalar inputs and outputs and therefore struggle to capture the detailed spatial behaviour of complex simulations. Field prediction is important because crash response depends not only on global metrics but also on the spatial distribution of deformation, intrusion, and load transfer across the structure. Convolutional neural networks (CNNs) have been used for field prediction by mapping simulation data to 2D or 3D image representations \cite{Kohar2021CAE,Li2024CNN,Ding2024FRPCNN,Attar2023SheetStampingDL,Attar2021HFQCNN}. Although these approaches can predict detailed fields, pixel- or voxel-based representations can struggle to encode complex geometries with irregular discretisations. To address these limitations, recent work has increasingly leveraged graph neural networks (GNNs) as surrogate models that directly encode vehicle components into graph representations \cite{Li2025ReGUNet,Wen2023SDNE}. Wen et al. \cite{Wen2023SDNE} used segment-based graphs to predict the dynamic behaviour of regularly structured vehicle components. Le Guennec et al. \cite{LeGuennec2025NeuralFieldsCrash} proposed a neural field surrogate model for crash dynamic prediction of vehicle components, showing superior performance over traditional reduced-order models. André et al. \cite{Andre2023MechanicalJointsCrash} combined neural networks with FE simulations to model mechanical joints in large-scale crash analyses, showing how learned component models can reduce the cost of full-vehicle simulation. Thel et al. \cite{Thel2024FEMIN,Thel2025FEMINCrash} introduced Finite Element Method Integrated Networks (FEMIN) as a framework that embeds neural networks within the FE pipeline to replace selected parts of the simulation while preserving physics-based structure. Zhang et al. \cite{Zhang2026MeshBasedGeometricDL} proposed a mesh-based GNN framework that compresses large multi-component FE assemblies into smaller graph representations for rapid response prediction. Nabian et al. \cite{Nabian2025CrashDynamicsML} applied MeshGraphNet \cite{Pfaff2021MeshGraphNets} and Transolver \cite{Wu2024Transolver,Luo2025TransolverPP} to a body-in-white crash dataset with more than 200 components using the NVIDIA PhysicsNeMo \cite{PhysicsNeMo2023}. These studies demonstrated the feasibility of GNN-based surrogate modelling for multi-component crash dynamics. 

GNN surrogate models are also developed for other mesh-based prediction in related domains \cite{Pfaff2021MeshGraphNets,Deshpande2022MAgNET,He2023DeepEnergy,Fu2023BoundaryGraphEmbedding,Chen2024DynamicDeformable,Zhou2025MultiLevelGCNN}. The most commonly adopted architecture is the encoder--processor--decoder architecture in the Graph Network-based Simulators (GNS) proposed by Sanchez-Gonzalez et al. \cite{SanchezGonzalez2020GNS}. This architecture applies multiple MLP-based graph-network blocks for iterative edge/node updates. MeshGraphNet (MGN) adapts this architecture to mesh simulation and augments mesh-edge interactions with additional world edges to better capture non-local contact and collision effects \cite{Pfaff2021MeshGraphNets}. To improve computational efficiency on large graphs, several multiscale models \cite{Fortunato2022MultiscaleMeshGraphNets,Cao2022BSMSGNN} perform message passing on hierarchies of coarsened graphs, reducing long-range message passing steps. 

MGN provides a strong foundation for modelling mesh data with message passing neural networks (MPNNs). This approach utilises MLPs as edge and node update functions, which are typically more transferable across different meshes with different topologies. The trainable update functions (weights) are shared across all nodes and edges within the graph, so we refer to this as shared-weight message passing. By contrast, Deshpande et al. \cite{Deshpande2022MAgNET} proposed the Multi-channel Aggregation Network (MAgNET), consisting of multi-channel aggregation layers that assign non-shareable, edge-specific weights to each edge in each channel. This can improve nonlinear approximation accuracy but requires a fixed or topology-consistent graph structure during training. As a result, compared with shared-weight message passing models, its application is limited when the input mesh topology changes.

\begin{figure}[t]
\centering
\begin{minipage}[t]{0.28\textwidth}
    \centering
    \includegraphics[width=0.9\linewidth]{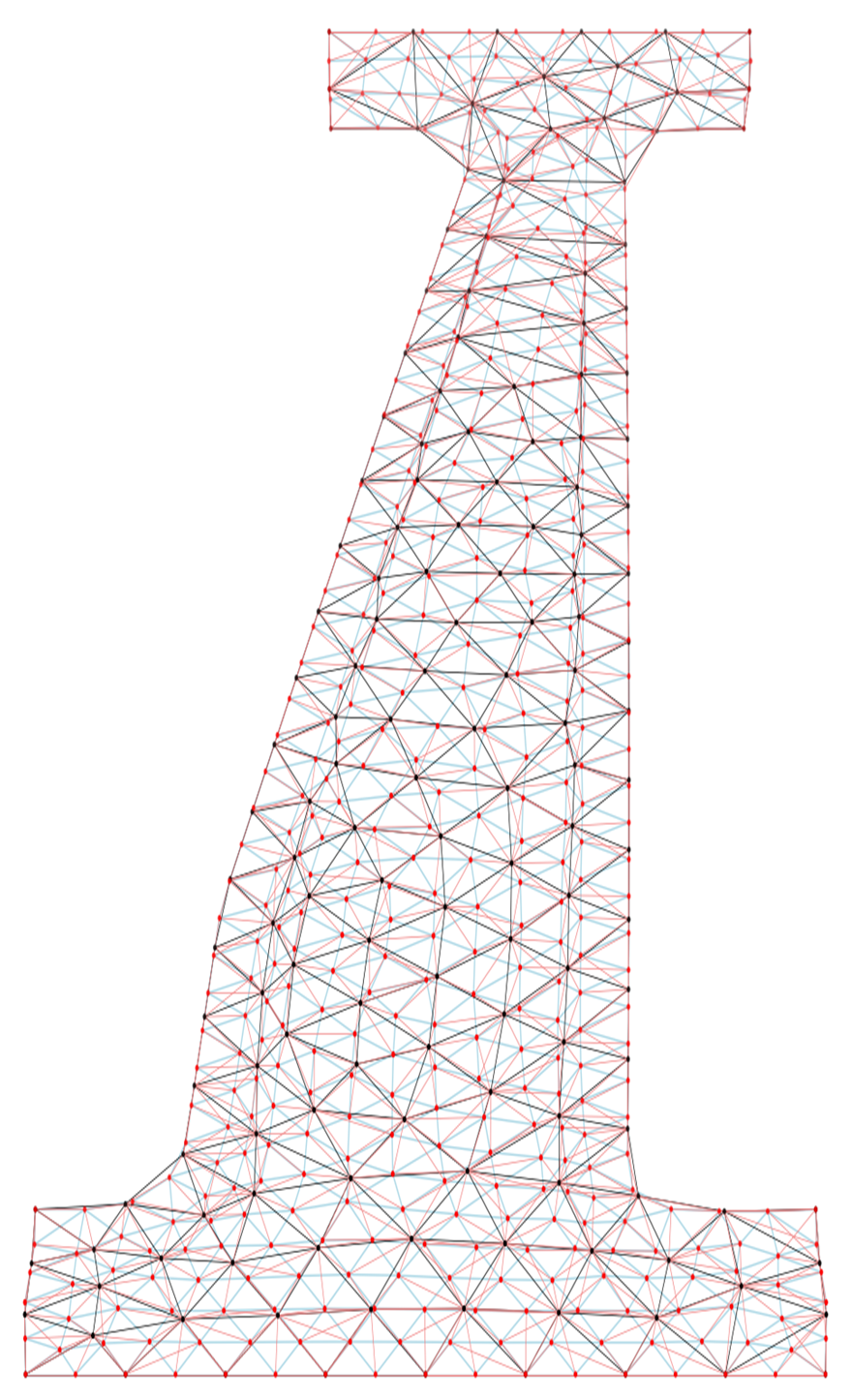}

    (a)
\end{minipage}
\hfill
\begin{minipage}[t]{0.52\textwidth}
    \centering
    \includegraphics[width=0.9\linewidth]{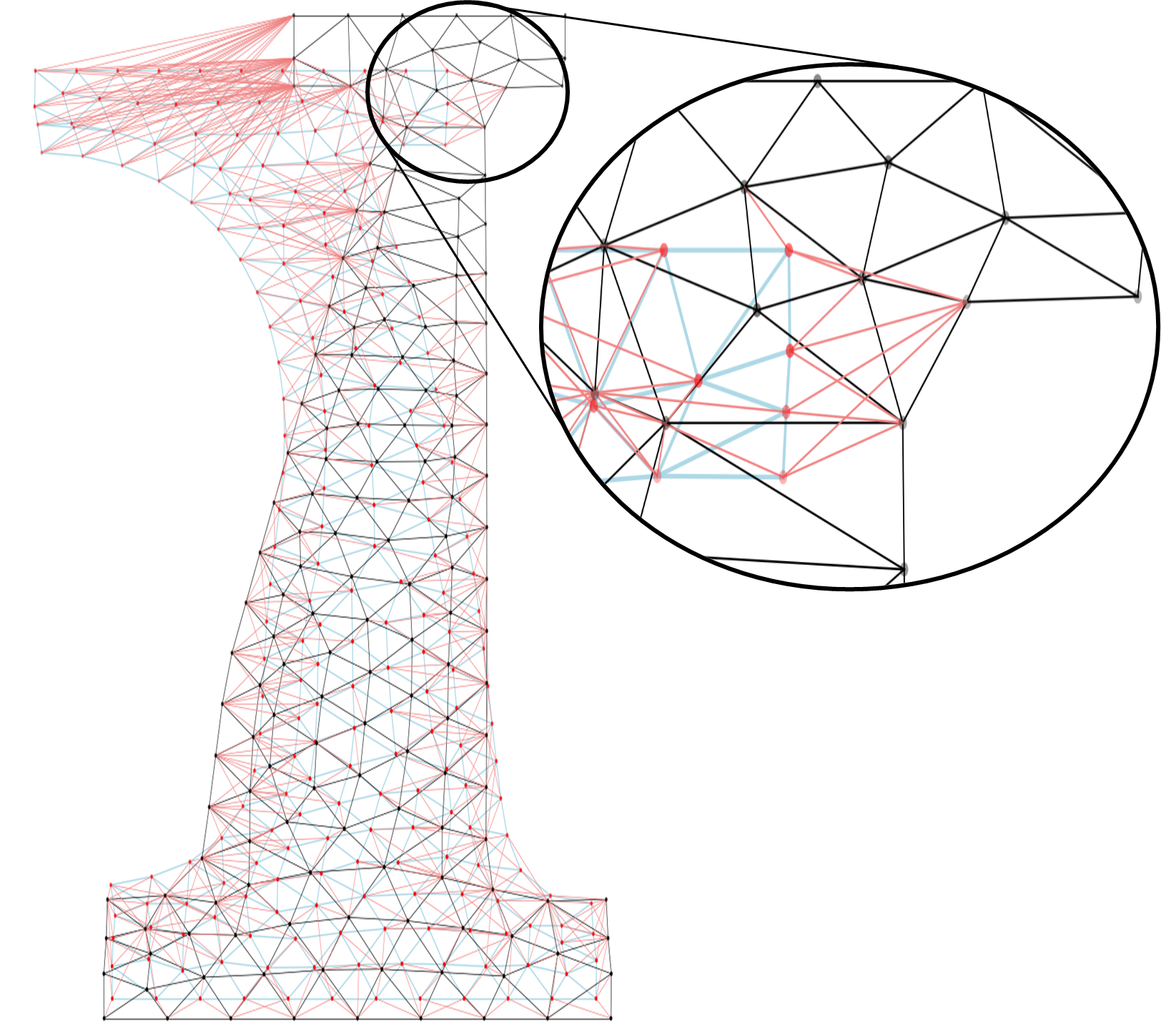}

    (b)
\end{minipage}

\vspace{0.5em}
{
\footnotesize
\setlength{\tabcolsep}{8pt}
\renewcommand{\arraystretch}{1.0}
\begin{tabular}{ccccc}
\textcolor{red}{$\bullet$} fine node & \raisebox{0.35ex}{\textcolor{cyan!40}{\rule{1.4em}{0.5pt}}} fine edge & \textcolor{black}{$\bullet$} coarse node & \raisebox{0.35ex}{\textcolor{black}{\rule{1.4em}{0.5pt}}} coarse edge & \raisebox{0.35ex}{\textcolor{red}{\rule{1.4em}{0.5pt}}} cross-graph edge
\end{tabular}
}
\caption{Illustration of spatial-proximity-based cross-graph edge construction. When the fine and coarse graphs have similar geometry, as in (a), nearest-neighbour fine-to-coarse edges provide meaningful local correspondence. Under large shape variation, as in (b), the fixed coarse graph becomes geometrically misaligned, causing some fine nodes to connect to inappropriate coarse regions.}
\label{fig:cross_graph_edge_connection}
\end{figure}

Prior mesh-based crashworthiness surrogates such as the Recurrent Graph U-Net (ReGUNet) \cite{Li2025ReGUNet,Zhao2026RUGNN} have shown that hierarchical Graph U-Net architectures with fixed coarsened graphs and edge-specific coarse-level operations can achieve accurate and efficient deformation prediction for vehicle panel components. However, the same mechanism that provides high prediction capacity also creates a generalisability bottleneck. Edge-specific layers require fixed coarse connectivity, while cross-graph edges are commonly constructed using spatial proximity between the input mesh and a shared coarse template. When geometric variation becomes large, spatial proximity alone can connect non-corresponding structural regions and degrade prediction performance. When the target distribution or component family changes more substantially, this limitation can also reduce transferability by increasing the amount of target data required for adaptation. Here, generalisability refers to prediction on unseen geometries without retraining or fine-tuning, whereas transferability refers to efficient adaptation using limited target data. As shown in Figure~\ref{fig:cross_graph_edge_connection}, spatial-proximity-based cross-graph connections can become insufficient for large shape variation, leaving many nodes unconnected and reducing predictive performance. This limitation reflects a fundamental trade-off: models with edge-specific operations can achieve higher predictive accuracy, but their reliance on fixed graph connectivity limits generalisability. This motivates a generalisable mesh-based surrogate that preserves fixed coarse topology for high-capacity edge-specific aggregation. Such a surrogate should also adapt the coarse graph geometry to each input shape and use transferable pretraining for efficient target adaptation.

In this paper, we address the trade-off between topological flexibility and predictive capacity in mesh-based GNN surrogates for crashworthiness analysis. The proposed Mask-Morph Graph U-Net retains fixed-topology coarsened graphs so that edge-specific multiscale aggregation layers can be used, but morphs the coarsened graph hierarchy to each input geometry before constructing fine-to-coarse cross-graph edges. This improves spatial correspondence under shape variation while preserving the trainable edge-specific structure at coarse levels. We further adopt a masked pretraining and parameter-efficient fine-tuning strategy to improve robustness, data efficiency, and transferability across tasks. The resulting framework is termed Mask-Morph Graph U-Net (MMGUNet). We evaluate MMGUNet on multiple crashworthiness scenarios, including B-pillar side-impact cases and a U-channel dynamic-loading case, and demonstrate improved predictive performance and cross-component transfer performance.

The main contributions of this work are as follows:
\begin{itemize}
    \item We propose Mask-Morph Graph U-Net, a multiscale mesh-based GNN surrogate that combines topology-preserving coarse-graph morphing with edge-specific downsampling and upsampling layers for crashworthiness field prediction.
    \item We introduce a feature-aligned barycentric morphing procedure that allows fixed-topology, edge-specific multiscale graph operators to be reused across geometrically varying finite-element meshes.
    \item We adopt a supervised masked pretraining and a parameter-efficient fine-tuning strategy for crashworthiness surrogate modelling to further improve generalisability and training efficiency.
    \item We construct and evaluate a multi-geometry crashworthiness case-study suite, including four B-pillar shape variants and one U-channel case, and provide comprehensive cross-task transfer learning results.
\end{itemize}

The remainder of this paper is organised as follows. Section~\ref{Related work} reviews related work. Section~\ref{Task and network} defines the task and presents the network architecture. Section~\ref{Dataset Generation} describes dataset generation and the case-study setup. Section~\ref{Morph} introduces feature-aligned morphing for shell meshes. Section~\ref{Training strategy} presents the training strategy, including masked pretraining and parameter-efficient fine-tuning. Section~\ref{Results and Discussion} reports and discusses the experimental results. Finally, Section~\ref{Conclusion} concludes the paper.

\section{Related work}
\label{Related work}

\subsection{Graph morphing} 
\label{Graph morphing}

For a generalisable fixed-topology graph surrogate, a key requirement is to maintain consistent coarse-level connectivity while improving geometric correspondence between fine and coarse graph levels. This can effectively improve the model's generalisability by constructing more meaningful cross-graph edge connections, while maintaining high predictive accuracy due to the edge-specific layers \cite{Li2025ReGUNet}. In this setting, the task can be formulated as template-to-target mesh morphing with fixed connectivity, specifically, updating nodal coordinates to follow the target surface, while preserving the source topology and avoiding remeshing.

In engineering practice, a commonly used morphing strategy is control-point-driven morphing \cite{deBoer2007RBF}. This approach defines a set of control points or control regions, prescribes translations at those locations, and then computes the displacements of the remaining nodes to ensure a smooth geometric transition. For example, given two meshes, if the objective is to morph a template mesh to match the shape of a target mesh, the boundary nodes of the template mesh can be constrained to coincide with the target mesh boundary. For interior nodes, a common choice is radial basis function (RBF) interpolation, which propagates boundary displacements to interior nodes and can handle large deformations without explicit connectivity dependence \cite{deBoer2007RBF}. Other widely used techniques include free-form deformation (FFD), which controls smooth global shape changes through a low-dimensional embedding lattice \cite{Sederberg1986FFD}. 

While these methods are efficient for geometry warping, they do not explicitly optimise surface-to-surface correspondence. Parameterisation-based morphing is a more effective approach in this context. Cross-parameterisation and inter-surface mapping methods provide a stronger foundation by directly seeking bijective maps between meshes \cite{Kraevoy2004CrossParameterization,Schreiner2004InterSurface}. A standard mesh-morphing pipeline is to map source and target surfaces to a common parameter domain, establish correspondence there, and then interpolate nodal coordinates to obtain the transformed shape \cite{Alexa2001MeshMorphing}. Traditional parameterisation approaches include barycentric embedding by Tutte \cite{Tutte1963}. In this formulation, the boundary is fixed as a convex polygon, and each interior vertex is placed at the barycentric average of its neighbouring vertices, which yields a crossing-free embedding under standard graph conditions. Because the interior vertex constraints are linear, the embedding can be computed efficiently as the unique solution of a sparse linear system for a given boundary. Later advancements, including Floater-style barycentric variants and mean-value coordinates, improve mapping quality while retaining the robustness of linear barycentric formulations \cite{Floater1997,Floater2003MVC}. Conformal methods such as least-squares conformal maps (LSCM) further reduce angular distortion \cite{Levy2002LSCM}. More recent injective optimisation frameworks such as scalable locally injective mappings \cite{Rabinovich2017SLIM} improve distortion control and reduce foldovers under larger shape variation.

\subsection{Masked pretraining and fine-tuning} 
\label{Mask review}

Another way to enhance generalisability is to use masked pretraining followed by task-specific fine-tuning. In computer vision, He et al. \cite{He2022MAE} proposed the masked autoencoder (MAE) which introduced high-ratio masking with asymmetric encoder--decoder design. The self-supervised pretraining procedure involves pretraining the encoder--decoder architecture with randomly masked input, followed by introducing masked tokens in the latent representation before reconstructing the output with a lightweight decoder \cite{He2022MAE,Devlin2019BERT}. In graph learning, masked graph autoencoder frameworks such as GraphMAE \cite{Hou2022GraphMAE} pretrain encoders by reconstructing masked node information, thereby improving the generalisability of the fine-tuned model on downstream tasks. A subsequent decoding-enhanced variant further improves transfer performance by strengthening the decoder design and reconstruction objective, which yields more informative latent representations for fine-tuning across tasks \cite{Hou2023GraphMAE2}. For mesh-based physics simulation applications, MeshMask extends this idea by masking a random subset of mesh nodes during pretraining. The pretrained encoder can then be fine-tuned on target simulation tasks with improved long-horizon accuracy and robustness \cite{Garnier2025MeshMask}. Garnier et al. \cite{Garnier2025MeshMask} have also examined the transfer learning ability of the proposed model and training strategy, indicating improved data efficiency when transferring a pretrained model to unseen tasks. More generally, the benefits of pretraining and fine-tuning have also been examined in the broader transfer-learning literature for GNN surrogates. Whalen et al. \cite{Whalen2022ReusableGNN} conducted a transfer-learning study on GNN-based surrogate models for trusses, showing that pretraining on similar datasets can effectively reduce data requirements and improve training efficiency when adapting to new tasks.

\section{Task definition and Network Architecture}
\label{Task and network}

\subsection{Task definition} 
\label{Task definition}

This study considers component-level crashworthiness field prediction under fixed loading, boundary, material, and contact settings. The geometric design of the component is varied, and the surrogate model predicts the terminal nodal displacement field after impact. This single-step formulation is appropriate for early-stage design studies in which the final deformed state and intrusion response are the primary quantities of interest. Under this controlled setup, the model's generalisation behaviour is evaluated by increasing the magnitude of shape variation, changing mesh density and component scale, and transferring across related component geometries.

We formulate the prediction problem as a single-step prediction task, similar to \cite{Li2024CNN,Deshpande2022MAgNET}. In this case, the input is the component shape and the output is the final-step deformation fields in x, y, and z. This formulation is well aligned with practical design optimisation, in which the terminal deformed state is typically sufficient for early-stage crashworthiness assessment and decision-making.

\subsection{Graph definition} 
\label{Graph definition}

We represent each FE mesh as a graph
\begin{equation}
\mathcal{G}=({V},{E}),
\end{equation}
where $V$ is the set of nodes and $E$ is the set of edges. The input node features are denoted by $\mathbf{V}=\{\mathbf{v}_i\}_{i=1}^{N_v}$, and the edge features are denoted by $\mathbf{E}=\{\mathbf{e}_k\}_{k=1}^{N_e}$. Here, $\mathbf{v}_i$ is the node feature vector associated with node $i$, $\mathbf{e}_k$ is the edge feature vector associated with edge $k$, and $N_v$ and $N_e$ denote the numbers of nodes and edges, respectively. Each graph node corresponds to a mesh node, and each graph edge corresponds to an element edge in the mesh.

A node $i$ has nodal coordinate $\mathbf{x}_{i}=[x_i,y_i,z_i]$. For an edge connecting nodes $(i,j)$, we define the edge features as the relative position vector as
\begin{equation}
\mathbf{e}_{ij}=\left[\mathbf{r}_{ij},\;\lVert \mathbf{r}_{ij} \rVert_2\right] \in \mathbb{R}^{4}, \mathbf{r}_{ij}=\mathbf{x}_i-\mathbf{x}_j,
\end{equation}
where $\lVert \mathbf{r}_{ij} \rVert_2$ denotes the Euclidean distance between nodes $i$ and $j$. For each undirected mesh edge, both directed edges $(i,j)$ and $(j,i)$ are included so that the signed relative-position vector is consistently defined during message passing.
These edge features are sufficient to encode mesh shape and local geometric relations. Therefore, absolute nodal coordinates are not required, and node features can be treated as optional positional encodings. The effect of different node feature designs is analysed in Section~\ref{Positional encoding}. The model's output is the predicted node-wise displacement:
\begin{equation}
\hat{\mathbf{y}}_i=[\Delta \hat{x}_i,\Delta \hat{y}_i,\Delta \hat{z}_i], \quad i=1,\dots,N_v.
\end{equation}

To improve message-passing efficiency on large meshes, we adopt a hierarchical graph representation by coarsening the fine graph to multiple lower-resolution levels, thereby reducing the number of nodes processed at deeper layers. To keep coarse-level topology constant, which is required for edge-specific layers, the coarsened graphs are constructed from software-generated template meshes and shared across samples \cite{Li2025ReGUNet}. Cross-graph message passing is then performed by connecting each fine node to its $k$ nearest coarse nodes. Because we use shared coarsened graphs, this connection method is only valid when the fine graph has a shape similar to the coarse graph. As shown in Figure~\ref{fig:cross_graph_edge_connection}, when the shapes differ substantially, fine nodes may be connected to inappropriate coarse nodes, thereby limiting model performance. In the zoomed view of Figure~\ref{fig:cross_graph_edge_connection}(b), the fine-mesh nodes located at the top-right corner are connected to coarse-mesh nodes in the T-joint region. Under a correct correspondence, they should instead connect to the top-right nodes of the coarse mesh. To address this limitation, we morph the coarsened graphs to match each input fine graph and construct cross-graph edges after morphing the coarse hierarchy. The morphing procedure is detailed in Section~\ref{Morph}.

\subsection{Network architecture} 
\label{Network architecture}
\begin{figure}[t]
    \centering
    \includegraphics[width=0.95\linewidth]{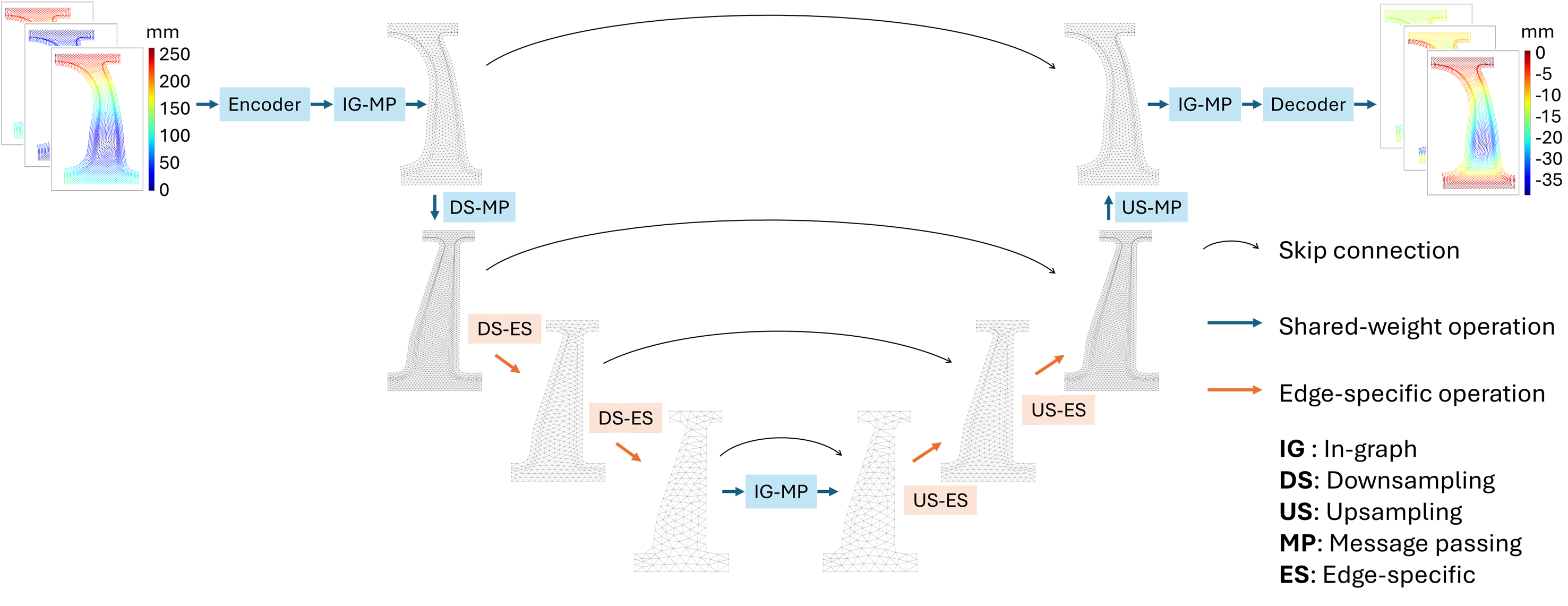}
    \caption{Overview of the proposed MMGUNet architecture. Shared-weight MLP-based operations are used for input encoding, in-graph message passing, and topology-flexible cross-graph operations, while edge-specific operations are used on fixed-topology coarsened graphs.}
    \label{fig:network_architecture}
\end{figure}

MMGUNet is built on a hierarchical Graph U-Net backbone for multiscale field prediction on FE meshes \cite{Li2025ReGUNet}. Unlike recurrent temporal graph surrogates designed for rollout prediction, the present architecture targets terminal crashworthiness field prediction under large geometric variation. As illustrated in Figure~\ref{fig:network_architecture}, the model adopts a hierarchical design based on the multi-level graph representation. Its main design objective is to combine topology-flexible MLP-based shared-weight operations at fine levels with high-capacity edge-specific operations on fixed coarsened graphs. The encoder maps node and edge inputs into latent states with MLP-based operators. At both the finest and coarsest levels, in-graph message passing (IG-MP) also uses MLP operators, which can be written as
\begin{equation}
\mathbf{m}_{ij}^{(l)}=\phi_e^{(l)}\!\left([\mathbf{h}_i^{(l)},\mathbf{h}_j^{(l)},\mathbf{e}_{ij}]\right),
\quad
\mathbf{h}_i^{(l+1)}=\phi_v^{(l)}\!\left([\mathbf{h}_i^{(l)},\sum_{j\in\mathcal{N}(i)}\mathbf{m}_{ij}^{(l)}]\right),
\end{equation}
where $\mathbf{m}_{ij}^{(l)}$ is the edge message for edge $\mathbf{e}_{ij}$, $\mathbf{h}_i^{(l)}$ is the latent feature for node $i$ at layer $l$, $\phi_e^{(l)}$ and $\phi_v^{(l)}$ are shared-weight MLP operators for edges and nodes, respectively, at layer $l$, and $\mathcal{N}(i)$ denotes all neighbouring nodes connected to node $i$.

A similar message passing layer is adopted for the first downsampling layer, i.e., from the input fine graph to the first shared coarsened graph. This layer enables the model to accommodate graphs with arbitrary topology because it can be applied to graphs with different numbers of edges. At coarser levels, we use edge-specific downsampling/upsampling layers (DS-ES and US-ES) on fixed-topology downsampled graphs \cite{Li2025ReGUNet}. The key idea is to assign edge-specific parameters per channel, thereby enabling more accurate approximation of nonlinear relationships. In an equivalent channel-wise form, the coarse-node update is

\begin{equation}
\mathbf{h}_{i}^{(l+1)}=\mathrm{LeakyReLU}\!\left(\sum_{j\in\mathcal{N}(i)}\sum_{c\in C_{in}}\mathbf{h}_{j}^{(c,l)}\cdot\mathbf{w}_{ij}^{(c,l)}\right),
\end{equation}
where $\mathbf{w}_{ij}^{(c,l)}\in\mathbb{R}^{C_{out}}$ is the $c$-th channel component of the edge-specific non-shareable weight for edge $e_{ij}$ at layer $l$, $C_{in}$ and $C_{out}$ are the numbers of input and output channels, respectively, and $\mathbf{h}_{j}^{(c,l)}$ is the channel-specific latent node feature for the $c$-th channel at layer $l$. Compared with shared-weight message passing, this edge-specific channel-wise parameterisation substantially increases representation capacity on fixed graphs, but it also increases trainable parameter count \cite{Li2025ReGUNet}.

The decoder maps final latent node states to the displacement prediction $\hat{\mathbf{y}}_i$ using MLP-based operators similar to those in the encoder.

\section{Dataset Generation}
\label{Dataset Generation}

This section describes the case studies constructed for this work. We focus on B-pillar-like thin-walled components under side-impact loading, and we design the dataset to systematically evaluate generalisability of the surrogate model to shape variations. In total, we consider seven case studies: six B-pillar cases and one U-channel case. 

\begin{figure}[t]
    \centering
    \begin{minipage}[t]{0.59\textwidth}
        \centering
        \includegraphics[width=0.95\linewidth]{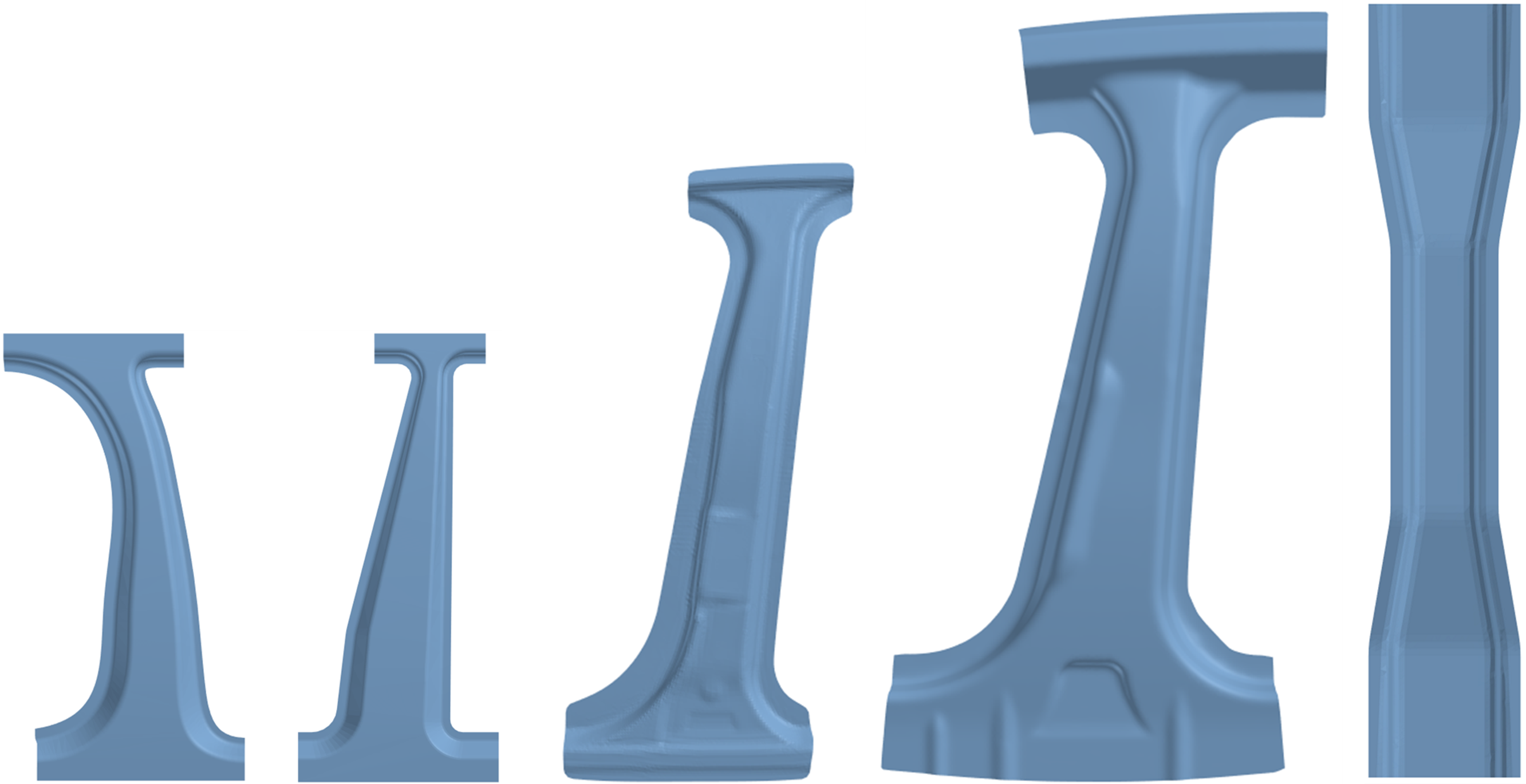}

        (a)
    \end{minipage}
    \hfill
    \begin{minipage}[t]{0.39\textwidth}
        \centering
        \includegraphics[width=0.95\linewidth]{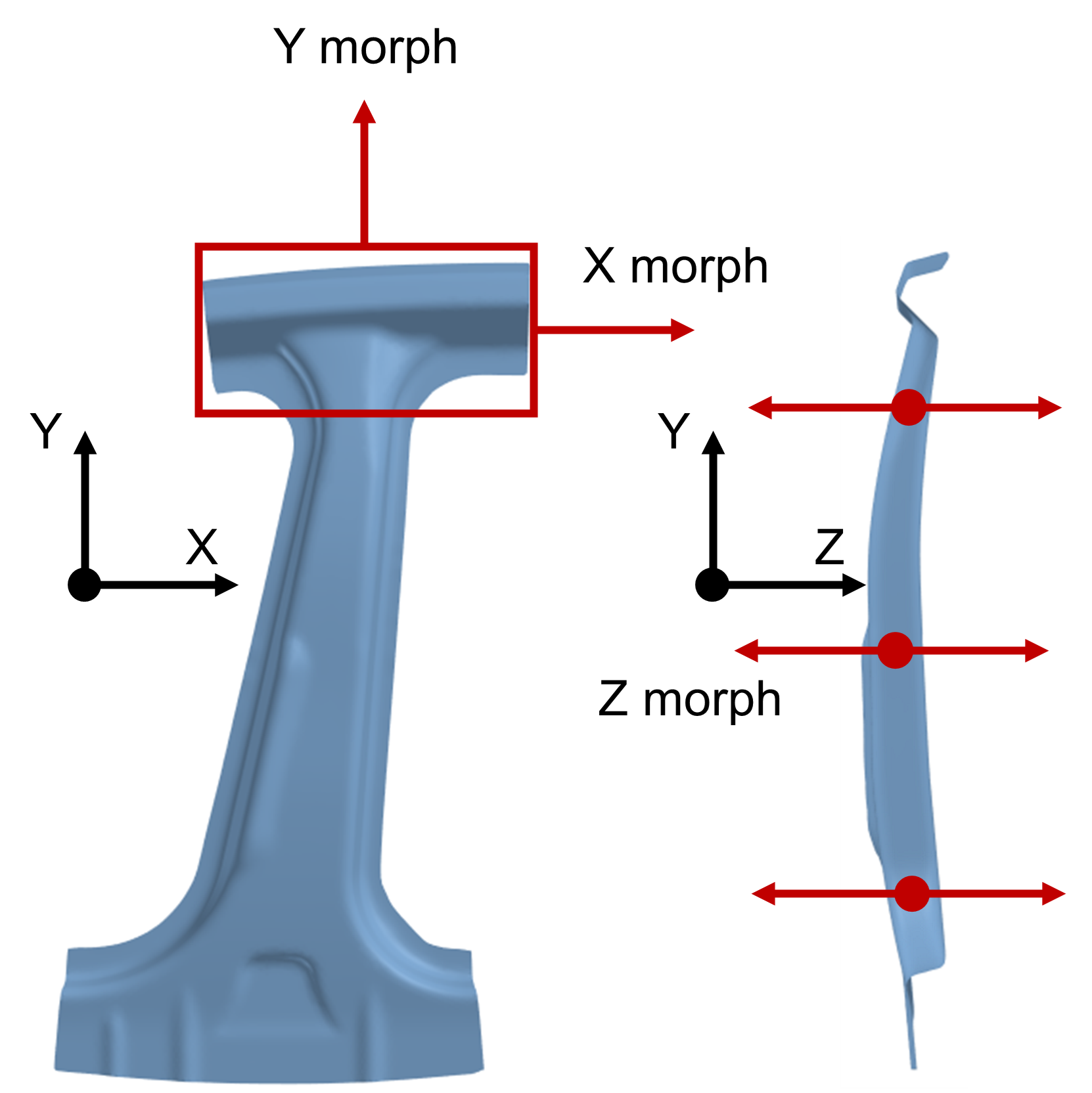}

        (b)
    \end{minipage}
    \caption{Case-study geometries and B-pillar morphing directions. (a) Base component geometries: B-pillar A, B-pillar B, B-pillar C, B-pillar D, and U-channel. (b) B-pillar design variation: top-region translations in the x- and y-directions and local morphing in the z-direction at one of three control points.}
    \label{fig:case_studies}
\end{figure}

\paragraph{B-pillar case studies}
We generate four base B-pillar shapes with increasing scale and geometric complexity. B-pillar A and B are approximately one-third of full scale. B-pillar C is a larger, more complex geometry at approximately two-thirds scale, and B-pillar D is full scale with the richest local geometric features. Figure~\ref{fig:case_studies}(a) shows that B-pillars C and D contain more complex local geometric features, such as stiffeners in the middle and wrinkles in the lower region. B-pillars A and B are created directly using computer-aided design (CAD) software, whereas B-pillars C and D are generated through hot-stamping simulations. As a result, B-pillars C and D provide more realistic component geometries and can include manufacturing-induced irregularities such as slight wrinkling. This design allows us to examine the model's ability to generalise across different component scales and levels of geometric complexity. 

\begin{figure}[t]
    \centering
    \begin{minipage}[t]{0.43\textwidth}
        \centering
        \includegraphics[width=0.95\linewidth]{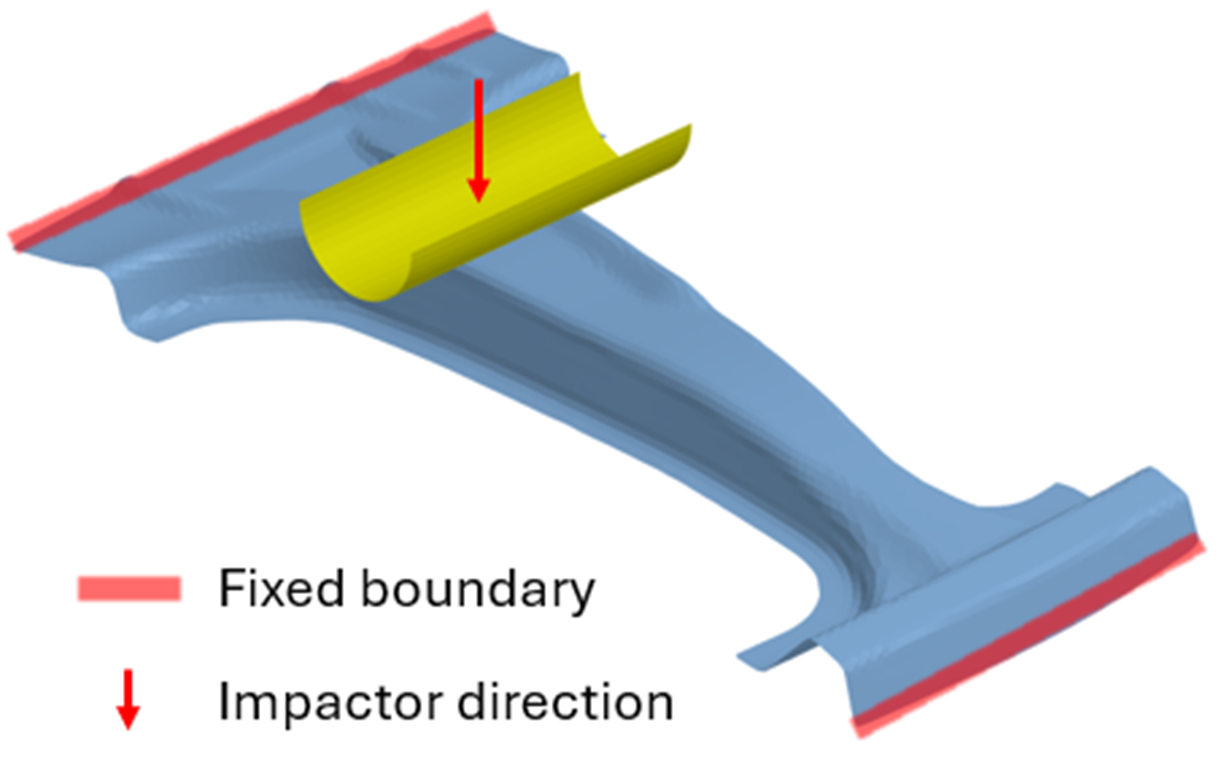}

        (a)
    \end{minipage}
    \hfill
    \begin{minipage}[t]{0.48\textwidth}
        \centering
        \includegraphics[width=0.95\linewidth]{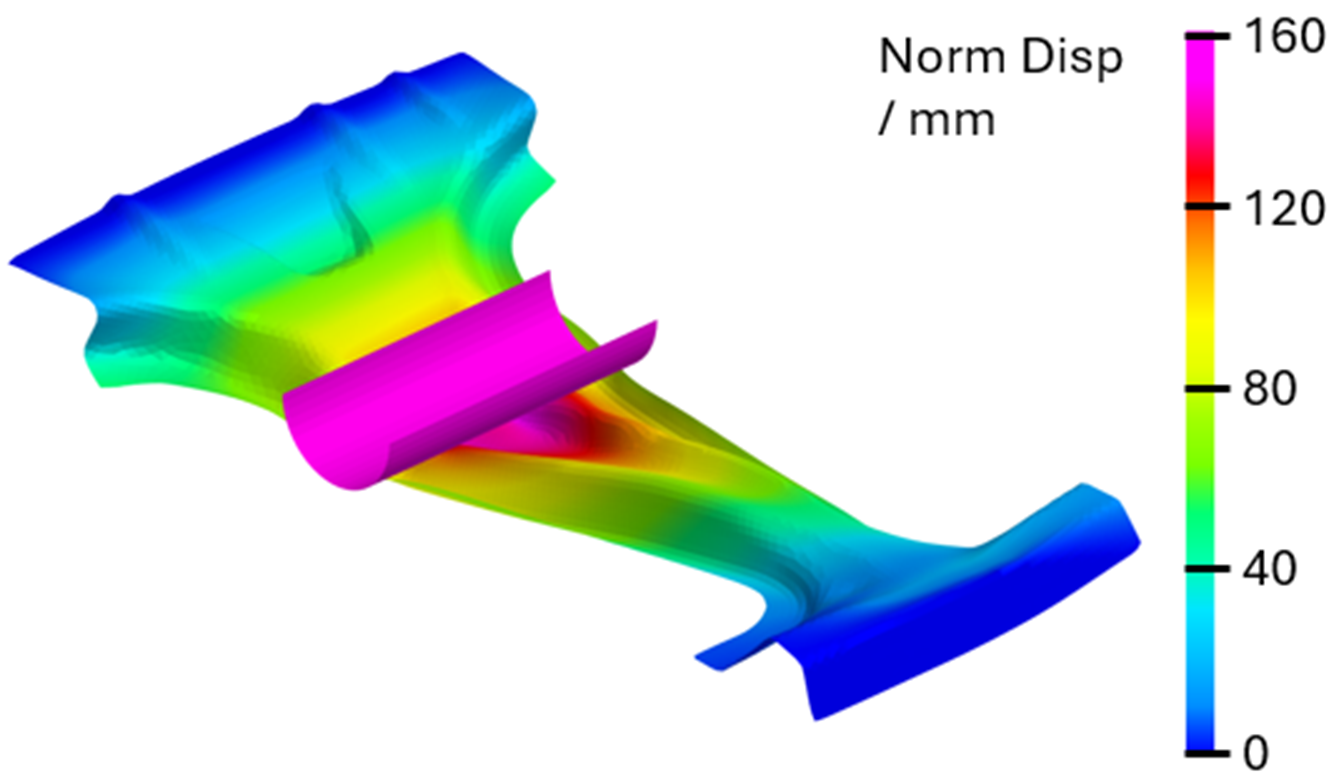}

        (b)
    \end{minipage}
    \caption{B-pillar side-impact simulation setup. (a) Fixed boundary regions and impactor direction. (b) Example deformed B-pillar response, coloured by nodal displacement norm in millimetres.}
    \label{fig:bpillar_conditions}
\end{figure}

For all B-pillar cases, boundary and loading conditions are fixed, as shown in Figure~\ref{fig:bpillar_conditions}. The FE setup simulates an experimental component-level B-pillar side crash test \cite{Zhang2020BPillarCollisionTest}, where the top and bottom regions are constrained in all degrees of freedom, and a cylindrical impactor strikes the component at the same location, located at one-third of the component height measured from the bottom, with a fixed impact velocity. FE simulations were run with Virtual Performance Solution; details can be found in \cite{Li2025ReGUNet}. Only geometry is varied, allowing us to isolate the effect of shape design on crashworthiness response. Shape variation is achieved by control point mesh morphing with Blender \cite{Blender2026}, followed by remeshing with HyperMesh \cite{AltairHyperMesh2026} to avoid mesh distortion and to create different mesh topologies for each sample. In the full setting, morphing is applied in all three directions (\(x,y,z\)). As illustrated in Figure~\ref{fig:case_studies}(b), the top region is translated in both the \(x\) and \(y\) directions. \(x\) is varied in \([-5\%, 5\%]\), \(y\) in \([0,10\%]\). The whole component is also morphed in \(z\) through one of three control points over the range \([-10\%,10\%]\). Appendix~A presents extreme morphed B-pillar cases to illustrate the extent of the geometric variation considered in this study. We sample the morphing parameters with Latin Hypercube Sampling (LHS) \cite{McKay1979LHS}. Beyond the four base B-pillar studies, we additionally define two reduced-direction variants of B-pillar A (B-pillar A1 for \(z\)-direction morphing only and B-pillar A2 for \(xz\)-direction morphing, while the full \(xyz\)-direction case is denoted B-pillar A3), resulting in six B-pillar case studies in total. These reduced variants are introduced to examine the model's ability to generalise across morphing directions, for example when trained on smaller directional variation and evaluated on larger variation. On the other hand, each B-pillar has a different mesh density as detailed in Table~\ref{tab:case_studies}. For instance, B-pillar B contains approximately twice as many mesh nodes as B-pillar A despite being defined at the same geometric scale. Examining these cases enables assessment of the model's ability to generalise across different mesh densities. Together, these B-pillar case studies allow us to systematically evaluate the model's generalisation behaviour with respect to shape, scale, mesh density, shape variation range, and local geometric complexity. The geometric details are summarised in Table~\ref{tab:case_studies}.

\paragraph{U-channel case study}
To further test cross-component generalisation, we include one U-channel case from the U-Channel-2 part-class of the U-Channel sheet metal (UCSM) dataset \cite{Lehrer2025UCSM}. The U-channel represents a stamped thin-walled sheet-metal component like the B-pillar, so the two cases share a similar manufacturing origin and basic structural characteristics. In addition, the U-channel retains a U-shaped profile similar to the middle region of a B-pillar, which makes the comparison interpretable while still involving a distinct overall component geometry. Together, these components are related enough to make cross-component transfer physically meaningful, yet different enough to provide a stringent test of how strongly the model generalises beyond the original component family. For the U-channel case, shape variation is generated by sampling CAD parameters rather than direct morphing. Specifically, we alter 16 geometric parameters with LHS \cite{McKay1979LHS}, including length, width, and height controls for the addendum, slant, and mid-plane regions. This leads to a broader and less constrained design space than the B-pillar cases. Details on the dataset parameters are provided in ~\ref{Dataset parameters}.

Table~\ref{tab:case_studies} summarises the seven case studies in terms of scale, approximate node count, variation definition, and geometric complexity. For each B-pillar case study, we generate 300 training samples and 50 test samples, with each set covering the full design space through an independent LHS design. For the U-channel case, owing to its larger design space, we use 1000 training samples and 200 test samples. The training and test sets are also sampled independently so that both provide coverage of the full design space.

\begin{table}[t]
    \centering
    \caption{Summary of the seven case studies used in this work.}
    \label{tab:case_studies}
    \setlength{\tabcolsep}{4pt}
    \renewcommand{\arraystretch}{1.1}
    \resizebox{\linewidth}{!}{%
    \begin{tabular}{lllll}
        \toprule
        Case study & Scale & Approximate node count & Shape variation range & Shape complexity \\
        \midrule
        B-pillar A1& $\sim$1/3 real size & ~3k & Morphing in $z$ only & Low \\
        B-pillar A2& $\sim$1/3 real size & ~3k & Morphing in $x$ and $z$ & Low \\
        B-pillar A3& $\sim$1/3 real size & ~3k & Morphing in $x,y,z$ & Low \\
        B-pillar B & $\sim$1/3 real size & ~6k & Morphing in $x,y,z$ & Medium \\
        B-pillar C & $\sim$2/3 real size & ~7k & Morphing in $x,y,z$ & High \\
        B-pillar D & Full scale          & ~8k & Morphing in $x,y,z$ & High \\
        U-channel  & Full scale          & ~5k & CAD-parameter variation & Medium \\
        \bottomrule
    \end{tabular}%
    }
\end{table}

\section{Parameterisation-based morphing with feature alignment}
\label{Morph}

The purpose of the morphing procedure is to improve fine-to-coarse graph correspondence while preserving the fixed topology required by edge-specific coarse-level layers. For each input mesh, the coarsened template graph is geometrically morphed toward the input shape before nearest-neighbour cross-graph edges are constructed. This allows the same coarse connectivity and edge-specific parameters to be retained, while reducing the risk that fine nodes are connected to geometrically inappropriate coarse regions.

In this section, we introduce the morphing algorithm adopted in this study to improve geometric shape alignment between fixed-topology hierarchical downsampled meshes and varying input fine meshes. The method first computes a barycentric embedding of each mesh onto a shared parameter domain (UV domain). The template mesh is then morphed by transferring the target 3D nodal coordinates through linear interpolation in the UV domain. To improve feature alignment, we further anchor the UV boundary to a case-specific polygon defined by the detected corner nodes, replacing the traditional circular boundary \cite{Floater1997}.

\subsection{Barycentric embedding}
\label{Tutte}

The objective of this step is to map the template and target shell meshes to a shared 2D domain so that node correspondence can be established and the required nodal displacement for shape transformation can be calculated. We follow the classic idea of Tutte's barycentric embedding theorem \cite{Tutte1963}. This is because computer-aided engineering (CAE) meshes for vehicle components are typically more well-structured than arbitrary meshes in the computer vision domain. Therefore, advanced anti-distortion algorithms are not required for the meshes considered in this study. In practice, we partition each mesh into boundary and interior nodes. We then prescribe the boundary positions on a predefined convex shape, referred to here as the UV domain. The boundary nodes must be identified using a consistent ordering and orientation. Specifically, the boundary loop is defined in a clockwise direction, starting from the bottom-left node of the component. 

We then solve for interior node coordinates in the UV domain using uniform barycentric averaging of neighbouring nodes. Because interior node constraints are linear, the mapping is obtained by solving a sparse linear system, yielding a unique embedded configuration for each mesh under fixed boundary conditions. This formulation is also physically interpretable as the equilibrium state of a pinned spring system. Further algorithmic details are provided in ~\ref{appendix:morphing_algorithm}.

\subsection{Boundary construction and feature alignment}
\label{Boundary feature alignment}

Although interior UV coordinates are determined by the linear system, the global correspondence pattern of the UV map is largely determined by how the boundary vertices are placed in the UV domain, for example whether they are distributed uniformly or anchored non-uniformly to specific feature locations such as corners. This is important for morphing, because interpolation in UV space implicitly assumes that semantically similar regions (e.g., corners and flanges) occupy comparable UV locations across different shapes. A generic boundary can introduce phase ambiguity, whereas feature-anchored boundaries help stabilise correspondence \cite{Kraevoy2004CrossParameterization,Schreiner2004InterSurface}.

A common choice is to place boundary vertices uniformly on a circular disk. This guarantees convexity and is easy to apply across diverse geometries, which is why it is widely used in standard applications of barycentric embedding \cite{Floater1997}. In this work, the circular boundary is treated as a baseline for morphing quality comparison.

For vehicle component shell meshes, boundary geometry usually contains a set of detectable landmarks. For example, we can detect corner nodes of a typical B-pillar or U-channel component. Therefore, instead of relying only on a circular boundary, we exploit these landmarks to improve feature alignment:
\begin{itemize}
    \item \textbf{Square option (4 anchors):} when four principal corners are detected, the boundary loop is mapped to a square by assigning these corners to the square vertices, while intermediate boundary nodes are distributed along each side according to boundary loop order.
    \item \textbf{Octagon option (8 anchors):} when eight corners are available (such as in B-pillar components), the boundary loop is mapped to a convex octagon. The eight detected corners are assigned to octagon vertices, and the remaining boundary nodes are linearly distributed along each octagon edge.
\end{itemize}

\begin{figure}[t]
    \centering
    \begin{minipage}[c]{0.34\linewidth}
        \centering
        \includegraphics[width=0.95\linewidth]{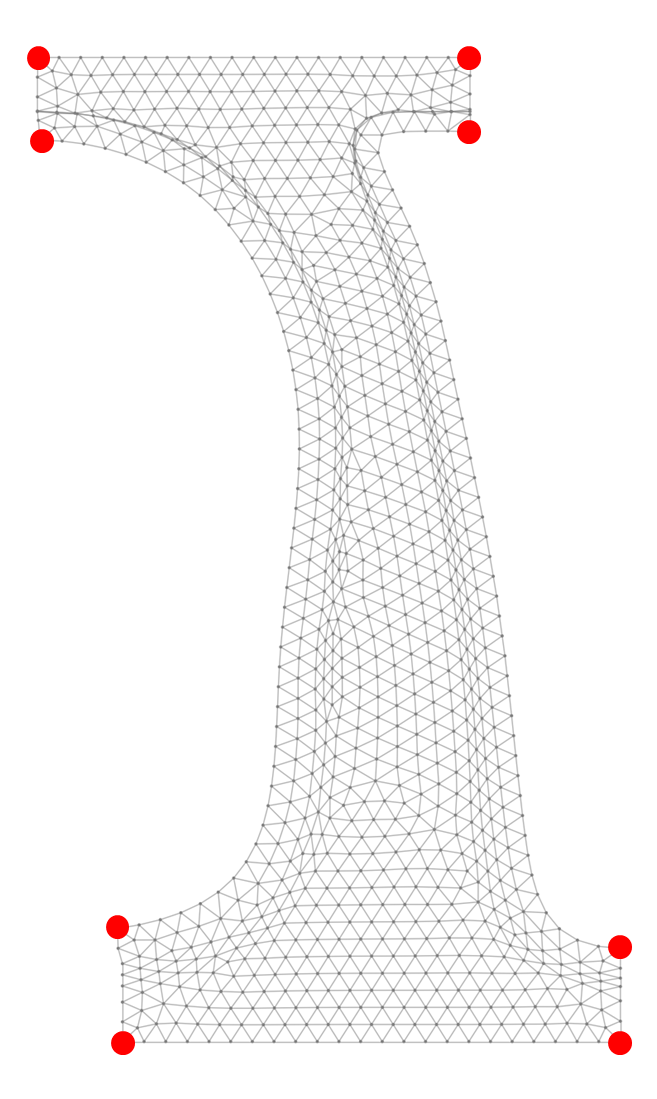}
        \par\vspace{1mm}
        {\small (a) B-pillar graph with detected corner nodes}
    \end{minipage}
    \hfill
    \begin{minipage}[c]{0.10\linewidth}
        \centering
        {\large $\rightarrow$}\\[22mm]
        {\large $\rightarrow$}\\[22mm]
        {\large $\rightarrow$}
    \end{minipage}
    \hfill
    \begin{minipage}[c]{0.50\linewidth}
        \centering
        \includegraphics[width=0.33\linewidth]{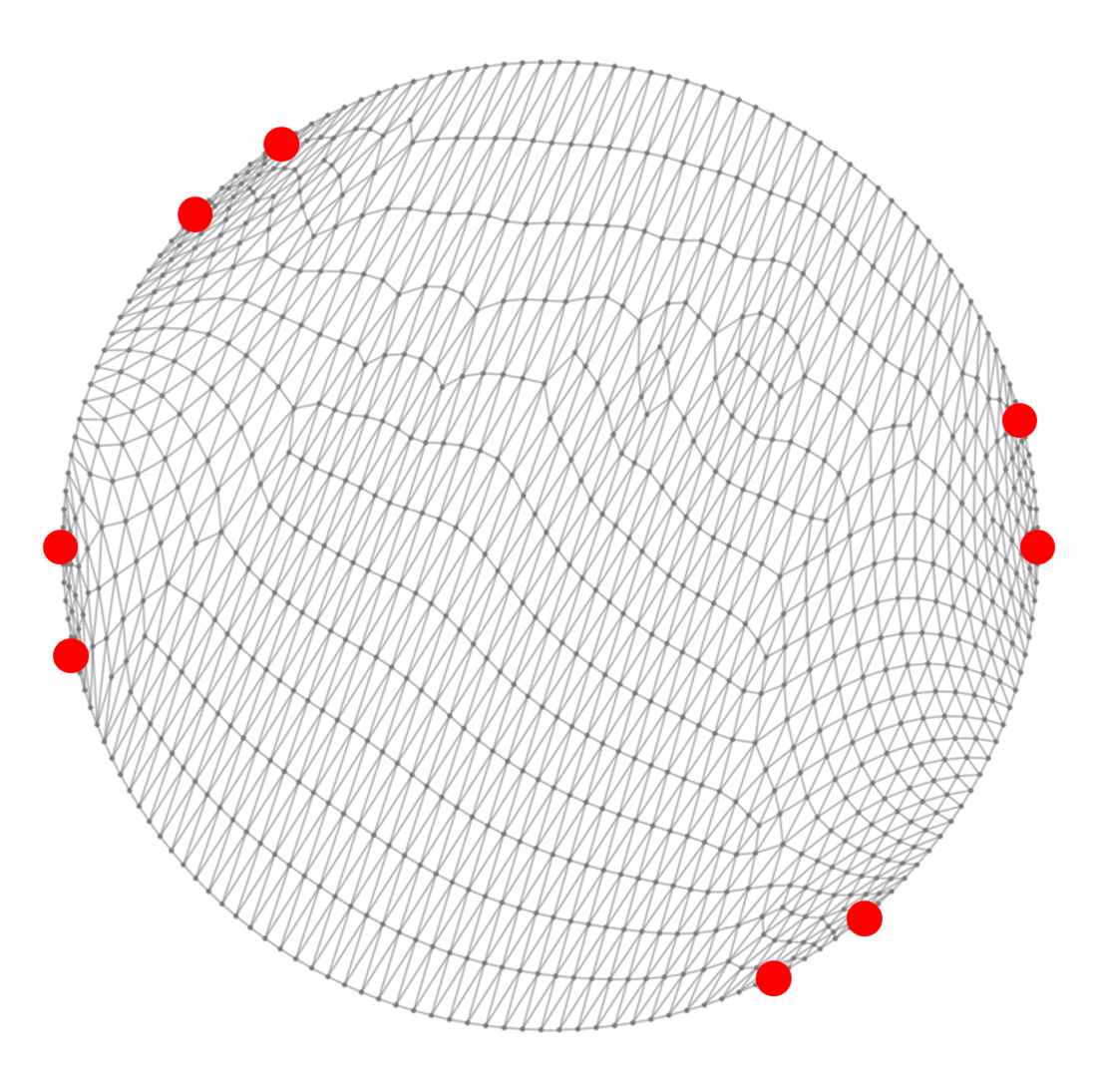}\\[1mm]
        {\small (b) UV mapping onto unit disk}\\[3mm]
        \includegraphics[width=0.33\linewidth]{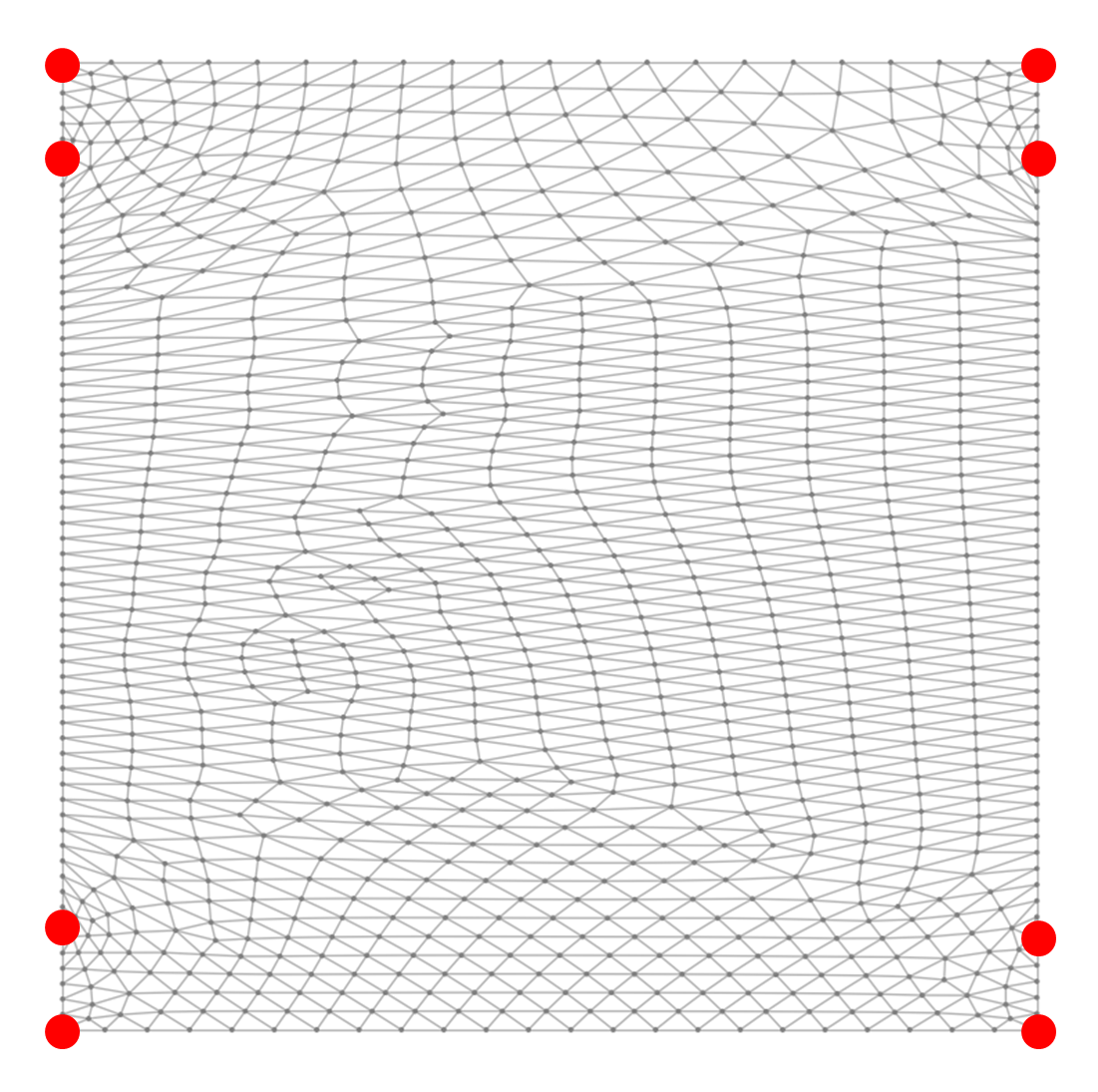}\\[1mm]
        {\small (c) UV mapping onto square domain}\\[3mm]
        \includegraphics[width=0.33\linewidth]{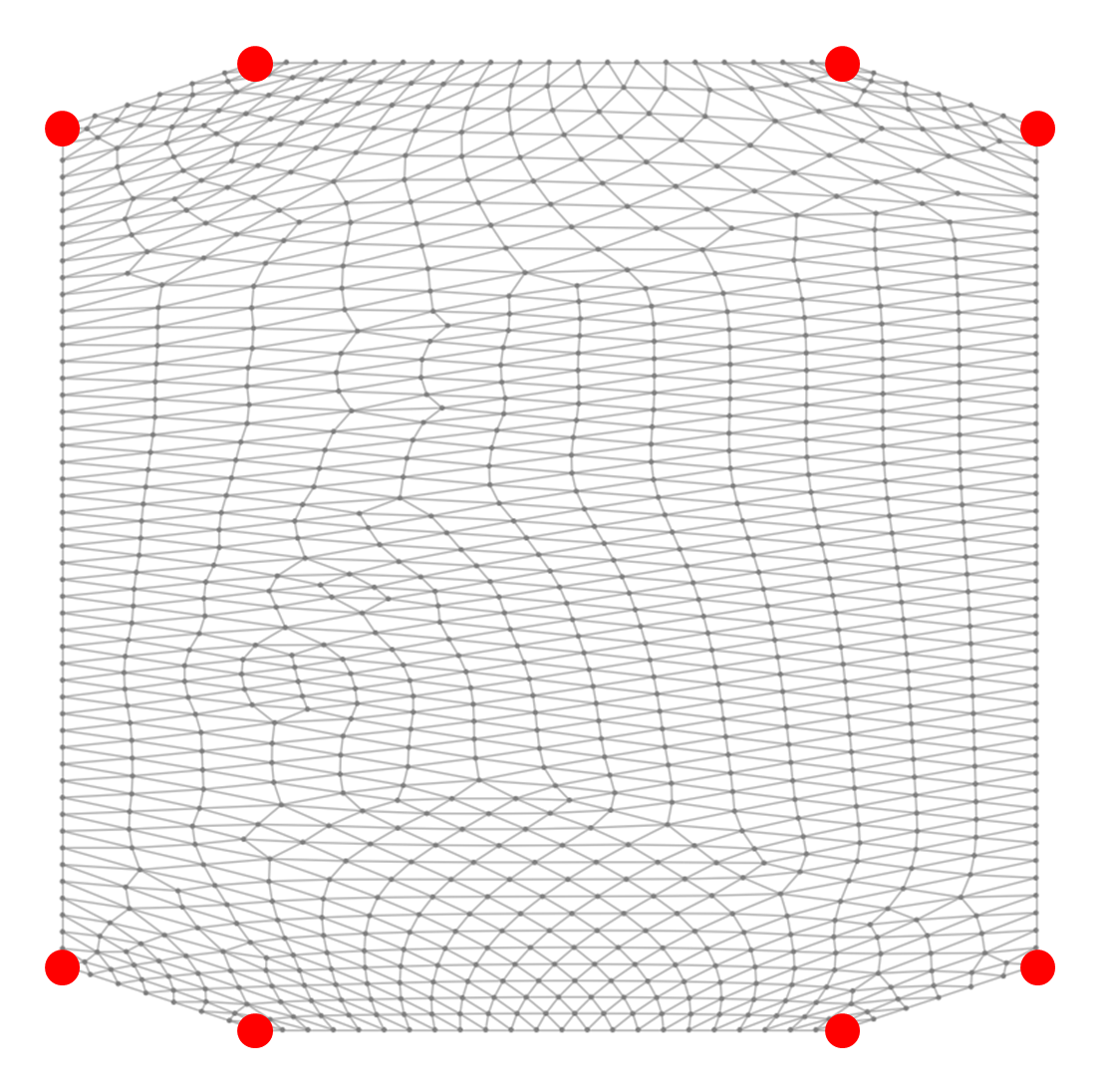}\\[1mm]
        {\small (d) UV mapping onto octagonal domain}
    \end{minipage}
    \caption{Feature-aligned UV parameterisation of a representative B-pillar mesh. (a) Graph of the representative B-pillar, where detected corner nodes are highlighted in red. (b)\textendash(d) Corresponding UV-domain graphs mapped onto different convex boundary domains: unit disk, square, and octagon, respectively.}
    \label{fig:bpillar_uv_mapping}
\end{figure}

In this study, we use the octagon boundary for the B-pillar case study and the square boundary for the U-channel case study. This geometric alignment step improves UV-domain consistency for downstream interpolation \cite{Kraevoy2004CrossParameterization,Schreiner2004InterSurface}. Figure~\ref{fig:bpillar_uv_mapping} illustrates the three boundary embeddings (disk, square, and octagon).

\subsection{Shape morphing via UV domain interpolation}
\label{Morphing and interpolation}
After mapping the template and target meshes to a common UV domain, we establish node-wise correspondence in UV space and compute the new 3D position of each template node by linear interpolation from the coordinates of the corresponding nodes on the target mesh. In other words, we define the updated nodal coordinates of each node of the template mesh based on interpolation from the coordinates from the target mesh. This keeps mesh connectivity unchanged, so topology is preserved throughout morphing. 

Let $\Phi_{\mathrm{temp}}:V_{\mathrm{temp}}\rightarrow\mathbb{R}^2$ denote the parameterisation map of the template mesh, which assigns each template node to its UV coordinate in the common parameter domain. For template node $i$, its UV coordinate is therefore
\begin{equation}
\boldsymbol{s}_i^{\mathrm{temp}}=\Phi_{\mathrm{temp}}(i).
\end{equation}
Next, let $T(\boldsymbol{s}_i^{\mathrm{temp}})$ denote the target UV element that contains the point $\boldsymbol{s}_i^{\mathrm{temp}}$. If the vertices of this target-space element are indexed by $q$ and the barycentric weights of $\boldsymbol{s}_i^{\mathrm{temp}}$ with respect to that element are $\lambda_q$, the corresponding target-space coordinate is obtained by barycentric interpolation:
\begin{equation}
\tilde{\mathbf{x}}_i^{\mathrm{tgt}}=\sum_{q\in T(\boldsymbol{s}_i^{\mathrm{temp}})} \lambda_q\, \mathbf{x}_q^{\mathrm{tgt}}.
\end{equation}
This makes the interpolation step explicit: $\Phi_{\mathrm{temp}}$ maps the template node into the common UV domain, $T(\cdot)$ identifies the enclosing target UV element, and the barycentric weights then interpolate the 3D target coordinates of that element.

\begin{figure}[t]
    \centering
    \begin{minipage}[t]{0.98\linewidth}
        \centering
        \includegraphics[width=\linewidth]{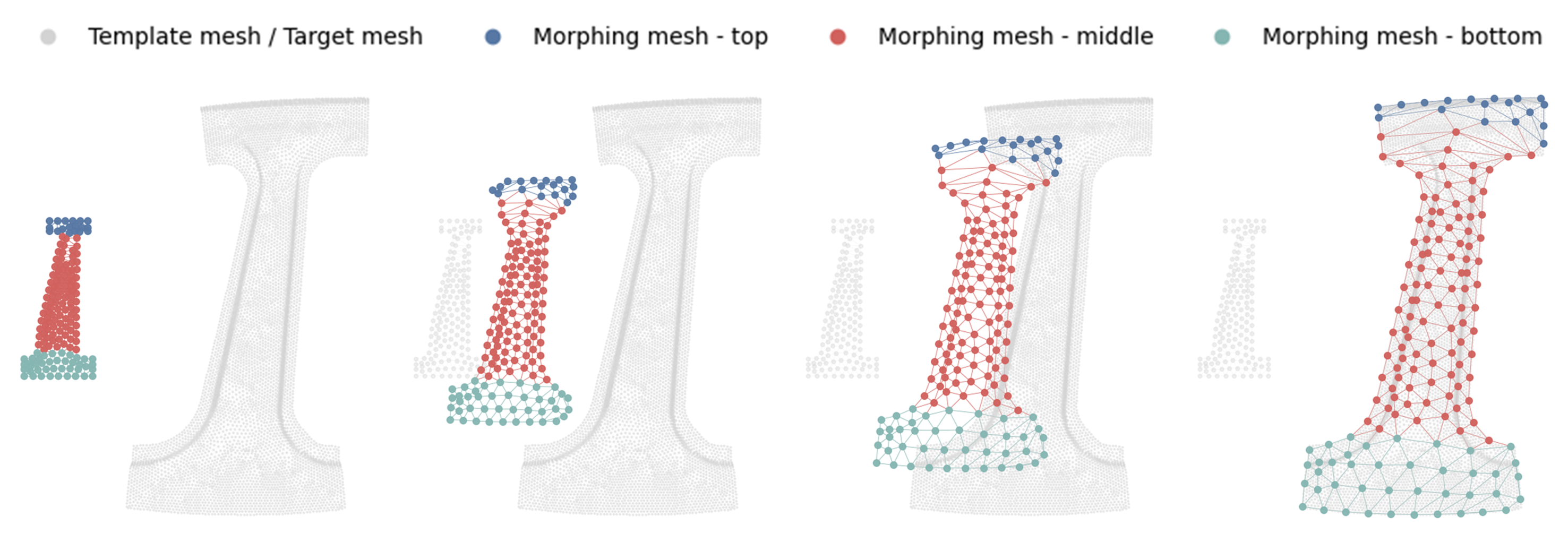}\\[1mm]
        {\small (a)}
    \end{minipage}\\[2mm]
    \begin{minipage}[t]{0.98\linewidth}
        \centering
        \includegraphics[width=\linewidth]{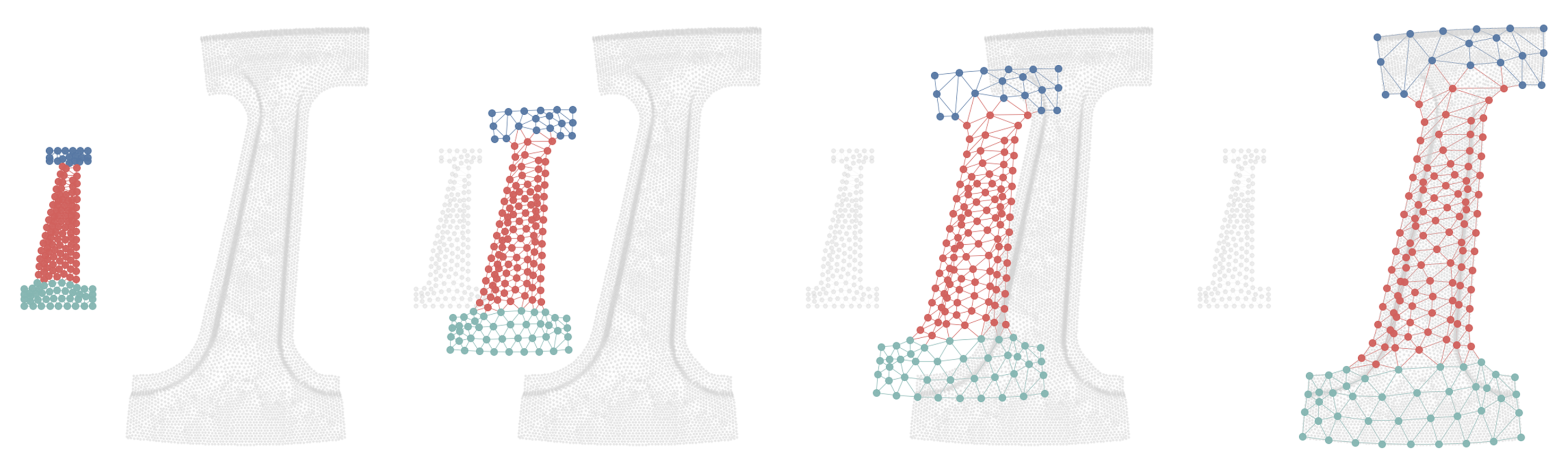}\\[1mm]
        {\small (b)}
    \end{minipage}\\[1mm]
    \caption{Four-step visualisation of template-to-target mesh morphing in a shared UV domain using (a) circular boundary mapping and (b) octagonal feature-aligned boundary mapping. The circular mapping exhibits a visible phase mismatch between corresponding regions, whereas the octagonal mapping preserves feature alignment and produces smoother template-to-target transitions.}
    \label{fig:morphing_circle_vs_octagon}
\end{figure}

The morphing can be visualised by calculating the intermediate morphing steps. Let $\mathbf{x}_i^{\mathrm{temp}}\in\mathbb{R}^3$ and $\tilde{\mathbf{x}}_i^{\mathrm{tgt}}\in\mathbb{R}^3$ denote the corresponding coordinates of node $i$ on the original and morphed template meshes, respectively. For an intermediate interpolation parameter $\alpha\in[0,1]$, the morphed coordinate is defined as
\begin{equation}
\mathbf{x}_i(\alpha)=(1-\alpha)\mathbf{x}_i^{\mathrm{temp}}+\alpha\tilde{\mathbf{x}}_i^{\mathrm{tgt}},
\end{equation}
where $\alpha=0$ recovers the template and $\alpha=1$ gives the target interpolated template coordinate. In the visual comparison in Figure~\ref{fig:morphing_circle_vs_octagon}, we show four evenly spaced interpolation states from the smaller B-pillar A configuration to the larger B-pillar D configuration. These states are obtained by sampling $\alpha$ uniformly from 0 to 1 to illustrate the full transition. We compare the four-step morphing trajectories obtained from circular and octagonal boundary parameterisation. For clearer visualisation, different regions of the component are labelled with different colours. From the colour distribution, clear distortion of the morphed shape under circular mapping can be observed. This indicates inconsistent feature alignment between template and target and leads to visibly distorted morphed shape. This can cause coarse nodes from different regions to connect to inappropriate fine-level nodes during cross-graph edge construction, therefore reducing message-passing efficiency. By contrast, octagonal mapping provides feature-anchored boundary alignment through consistent corner indexing, which removes the arbitrary phase ambiguity from the circular domain. As a result, nodes are morphed to their correct target locations (e.g., top region nodes remain in top regions), so cross-edge connections constructed from spatial proximity are physically meaningful and support effective message passing.

\section{Training strategy}
\label{Training strategy}

We train the model in two stages: masked pretraining followed by parameter-efficient fine-tuning. In the masked pretraining stage, a subset of nodes and their associated edges is randomly masked, and the model is trained to predict the full response from the partially observed graph. In the subsequent fine-tuning stage, the pretrained model is adapted to the target task while updating only a restricted subset of parameters. The motivation for combining these two stages is twofold. First, models with complex hierarchical architectures tend have an increased risk of overfitting \cite{Fortunato2022MultiscaleMeshGraphNets, Li2025ReGUNet}. Prior masked representation learning studies show that randomly masking nodes acts as a strong stochastic regulariser, improving generalisability and transferability \cite{Hou2022GraphMAE,Hou2023GraphMAE2}. Second, the edge-specific downsampling layers introduce a large number of trainable parameters and therefore high training cost. Masking reduces the number of active nodes and edges processed per iteration, which lowers computation and accelerates the training procedure. The parameter-efficient fine-tuning policy further limits overfitting and enables more efficient training.

\subsection{Masked pretraining}
\label{Masked pretraining}
In contrast to masked graph autoencoder frameworks and MeshMask-style pretraining pipelines that optimise separate encoder and decoder networks, we train a single architecture throughout both masked pretraining and fine-tuning. We pretrain the model in a supervised manner, where the model's output is the full displacement field.

In each training iteration, we apply a random mask to a subset of graph nodes and remove their associated edges from message passing. The masking is propagated across the first downsampling layer. When a fine-level node is masked, the associated fine-graph edges are masked, and the corresponding fine-to-coarse cross-graph edges are also masked. This design follows the general pipeline used in prior masked graph pretraining studies \cite{Hou2022GraphMAE,Garnier2025MeshMask}, while being adapted to our hierarchical graph setting. However, masked nodes remain in the output graph after graph upsampling and the supervised loss is computed on all nodes. 

To preserve physically critical constraints, we define a protected node set that is never masked. This set includes (i) nodes constrained by boundary conditions and (ii) nodes in the contact region with the impactor. Therefore, masking is applied only to the nodes whose motion is not directly prescribed by supports or immediate contact constraints. This constrained masking policy prevents the model from discarding key boundary or contact information that governs the global deformation response.

The mask ratio is a predefined hyperparameter and is fixed during training for a given experiment. Ablation results and sensitivity analysis with respect to mask ratio are provided in Section~\ref{Mask ratio}.

The term pretraining in this work is not restricted to any single case study or downstream task. Instead, the model can be pretrained on any set of structurally similar case studies and then fine-tuned for the target application. This includes a single-case setting, in which both pretraining and fine-tuning are conducted on the same case study, as well as a cross-case setting, in which the model is pretrained on one or multiple related case studies and subsequently fine-tuned on other unseen cases. In this sense, pretraining is used as a general representation-learning stage that is decoupled from a specific evaluation task. More detailed definitions of the cross-case protocols are provided in Section~\ref{Generalisability study training strategy}.

\subsection{Parameter-efficient fine-tuning}
\label{Parameter-efficient fine-tuning}

After pretraining, the model is fine-tuned on unmasked graphs using a parameter-efficient update policy. Specifically, we freeze all edge-specific downsampling and upsampling layers, as well as the coarse-level IG-MP layer, and update only the remaining modules. The main rationale is that the edge-specific downsampling and upsampling layers primarily encode how information is transferred across graph levels in the hierarchy. Because these layers operate on the shared coarsened graph topology, the learned inter-level propagation pattern is reusable across components that follow the same downsampling strategy. 

We also freeze the coarse-level IG-MP layer because the considered components belong to the same structural family and are subject to the same boundary conditions. Under this setting, the coarse-scale relationship between input geometry and output deformation is expected to remain largely consistent across cases. By contrast, the main source of variation lies in the input shape, which is encoded mainly through the encoder, the first fine-level IG-MP layer, and the DS-MP layer. These parts are therefore left trainable during fine-tuning so that the pretrained model can adapt to new geometries while preserving reusable multiscale propagation patterns learned during pretraining.

This policy substantially reduces the number of trainable parameters. In an example architecture used in this work, the full model contains 37.02M parameters in total, of which only 0.18M remain trainable during fine-tuning, corresponding to approximately 0.48\% of the full parameter count. As a result, fine-tuning is more efficient in both training time and GPU memory consumption while still allowing task-specific adaptation.

\subsection{Training details}
\label{Training details}

For model hyperparameters, unless otherwise stated, we use a batch size of $4$ and a learning rate of $4\times10^{-4}$, which were found to provide the best training performance in our preliminary tests. The encoded fine-level hidden channel dimension is set to $32$, and each DS-ES layer doubles the channel dimension so that the coarse-level hidden channel dimension reaches $128$. We use two fine-level message-passing steps and $15$ coarse-level message-passing steps.

The model outputs node-wise 3D displacement fields after impact. Training is supervised using mean squared error (MSE) over all nodal displacement components. For evaluation, we report two metrics. The first is the mean nodal Euclidean distance (MED), measuring how well the general deformed shape is predicted:
\begin{equation}
\mathrm{MED}=\frac{1}{N_v}\sum_{i=1}^{N_v}\lVert \hat{\mathbf{y}}_i-\mathbf{y}_i \rVert_2.
\end{equation}
where $\mathbf{y}$ and $\hat{\mathbf{y}}$ denote ground-truth and predicted nodal displacement fields, respectively. The second is the percentage error of maximum intrusion (MIPE), which measures the relative error in the peak absolute intrusion magnitude along the $z$ direction:
\begin{equation}
I(\mathbf{y})=\max_i \left|y_{i,z}\right|,
\end{equation}
\begin{equation}
\mathrm{MIPE}(\%)=\frac{\left|I(\hat{\mathbf{y}})-I(\mathbf{y})\right|}{I(\mathbf{y})+\epsilon}\times 100\%.
\end{equation}
where $y_{i,z}$ denotes the $z$-component of the nodal displacement at node $i$, $I(\cdot)$ denotes the maximum absolute intrusion over all nodes, and $\epsilon$ is a small positive constant added for numerical stability. In this study, the z-axis is aligned with the intrusion direction used for evaluating the terminal displacement response.

\subsection{Generalisability study training strategy}
\label{Generalisability study training strategy}

To evaluate different aspects of model generalisation, we train the model under four protocols using different combinations of case studies. The first protocol is \emph{single-case} training, in which the model is trained using only the target case. This setting evaluates model performance on a specific case study, all ablation studies and baseline comparisons are conducted under this protocol. The second protocol is \emph{all cases}, in which the model is trained using all available case studies, including the target case. This setting is used to assess whether additional training cases can improve predictive accuracy on the target case. The third protocol is \emph{all but target}, in which the model is trained using all available source cases except the target case. This provides a direct test of the model's ability to generalise to unseen out-of-distribution cases. The fourth protocol is \emph{transfer learning}, in which the model is first pretrained on the source cases and then fine-tuned using the target case only.

For the first three protocols, the models are trained for 2000 epochs. When masked pretraining and parameter-efficient fine-tuning are applied, even for these non-transfer protocols, the model is pretrained for 1000 epochs and then fine-tuned for an additional 1000 epochs using the same dataset split. For the transfer learning protocol, the models are pretrained for 1000 epochs on the source cases and then fine-tuned for 1000 epochs on the target case.

For the B-pillar studies, we define four trial groups to evaluate different generalisation regimes:
\begin{itemize}
    \item \textbf{Trial A (design space generalisation):} the target case is B-pillar A3, while the source cases include B-pillar A1 and A2. This evaluates generalisation to a wider range of design space (morphing range).
    \item \textbf{Trial B (shape generalisation):} the target case is B-pillar B, while the source cases include B-pillar A1, A2, and A3. These cases are at similar scale but differ in geometry, thereby testing generalisation to an unseen shape, as well as mesh density.
    \item \textbf{Trial C (scale/complexity generalisation-I):} the target case is B-pillar C, while the source cases include all other B-pillar cases. This evaluates generalisation to scale and geometric complexity differences.
    \item \textbf{Trial D (scale/complexity generalisation-II):} the target case is B-pillar D, while the source cases include all other B-pillar cases. This further evaluates generalisation to full-scale and the most geometrically complex case.
\end{itemize}

\begin{table}[t]
\centering
\caption{Summary of the trial settings used in the generalisability and transfer studies.}
\label{tab:trial_summary}
\small
\setlength{\tabcolsep}{4pt}
\begin{tabular}{p{0.15\textwidth} p{0.27\textwidth} p{0.26\textwidth} p{0.18\textwidth}}
\hline
\textbf{Trial name} & \textbf{Purpose} & \textbf{Source cases} & \textbf{Target cases} \\
\hline
Trial A & Design-space generalisation & B-pillar A1, A2 & B-pillar A3 \\
Trial B & Shape and mesh-density generalisation & B-pillar A* & B-pillar B \\
Trial C & Scale/complexity generalisation & B-pillar A*, B, D & B-pillar C \\
Trial D & Scale/complexity generalisation & B-pillar A*, B, C & B-pillar D \\
\hline
\end{tabular}
\\[0.25em]
\raggedright\footnotesize * B-pillar A includes A1, A2, and A3.
\end{table}

Table~\ref{tab:trial_summary} summarises all trial settings considered in this study. These trials are designed to evaluate the model's cross-case generalisability under different forms of distribution shift, and their quantitative results are presented in Section~7.3.

We additionally consider cross-component transfer learning using the U-channel case. In this setting, the model is pretrained on the B-pillar cases and then fine-tuned on the U-channel case, thereby evaluating transfer across component families. For this trial, the B-pillar and U-channel cases share a common coarse graph hierarchy, which means that the same coarse-level topology and associated edge-specific parameters are retained during transfer. The coarse-level node coordinates are morphed to the target U-channel geometry before cross-graph edges are constructed using the square UV domain during morphing. This is referred to as Trial E in later sections. 

Together, these settings provide a structured comparison of in-distribution training, out-of-distribution generalisation, and transfer learning across variation range, shape/mesh density, scale/complexity, and component type.

\section{Results and Discussion}
\label{Results and Discussion}

In this section, we evaluate the predictive performance and generalisation capability of the proposed method across the B-pillar and U-channel case studies. We first perform a series of ablation experiments to identify suitable design choices and hyperparameters for the proposed architecture and training strategy. We then compare MMGUNet with representative baseline methods on \textit{single-case} training. After that, we study the out-of-distribution and cross-case generalisation performance of MMGUNet. The main finding is that the proposed combination of morphing-based multiscale graph modelling, masked pretraining, and parameter-efficient fine-tuning improves accuracy while also strengthening generalisation to unseen shape variations and transfer settings.

\subsection{Ablation study}
\label{Ablation study}

\subsubsection{Positional encoding}
\label{Positional encoding}

As discussed in Section~\ref{Graph definition}, geometric information is primarily encoded through edge features, while node features are optional. In this ablation, we evaluate whether adding positional node features improves prediction quality. Performance is assessed with Morph-GUNet (MGUNet) without masking, on B-pillar A3 using MED and MIPE.

We compare four positional encoding components:
\begin{itemize}
    \item \textbf{Zeros:} node features are initialised to zeros (no explicit positional encoding).
    \item \textbf{One-hot (1H):} one-hot node-type encoding with three classes: boundary nodes (fixed at top/bottom boundaries), contact nodes (in contact with the impactor), and free nodes (all remaining nodes).
    \item \textbf{Laplacian eigenvector (LE):} the first 16 nontrivial Laplacian eigenvectors of the input graph \cite{Dwivedi2021GraphTransformer}.
    \item \textbf{Distance to critical regions (DTC):} Euclidean distances from each node to critical regions (to the closest fixed boundaries and impactor-contact region).
\end{itemize}

\begin{figure}[H]
    \centering
    \begin{minipage}[t]{0.98\linewidth}
        \centering
        \includegraphics[width=\linewidth]{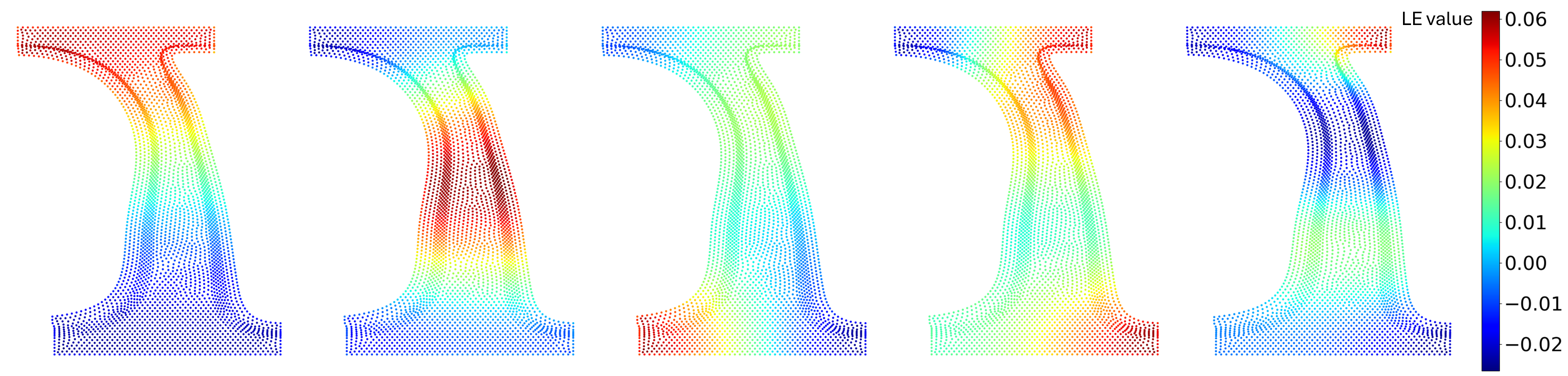}
        \\[1mm]
        {\small (a)}
    \end{minipage}\\[2mm]
    \begin{minipage}[t]{0.43\linewidth}
        \centering
        \includegraphics[width=\linewidth,height=34mm,keepaspectratio]{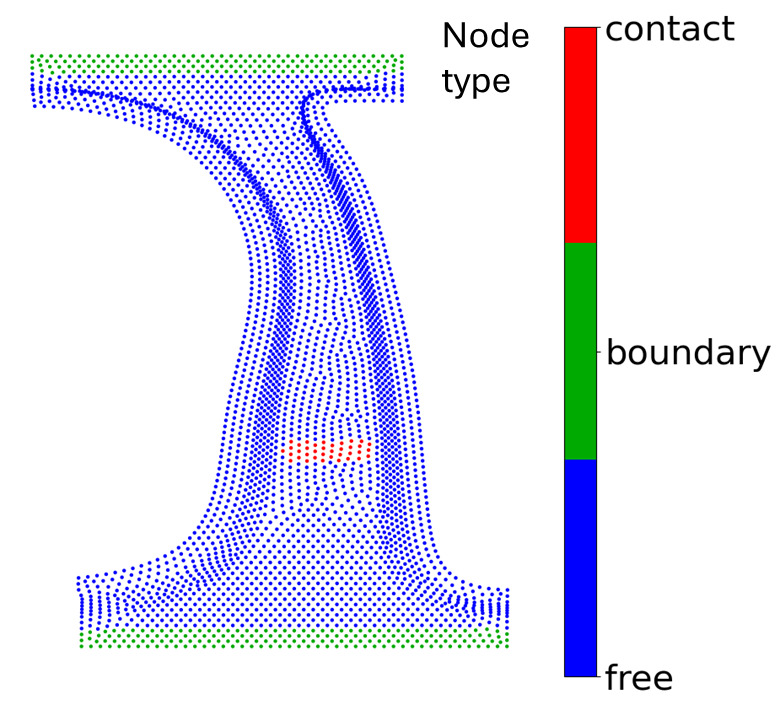}
        \\[1mm]
        {\small (b)}
    \end{minipage}\hfill
    \begin{minipage}[t]{0.55\linewidth}
        \centering
        \includegraphics[width=\linewidth,height=34mm,keepaspectratio]{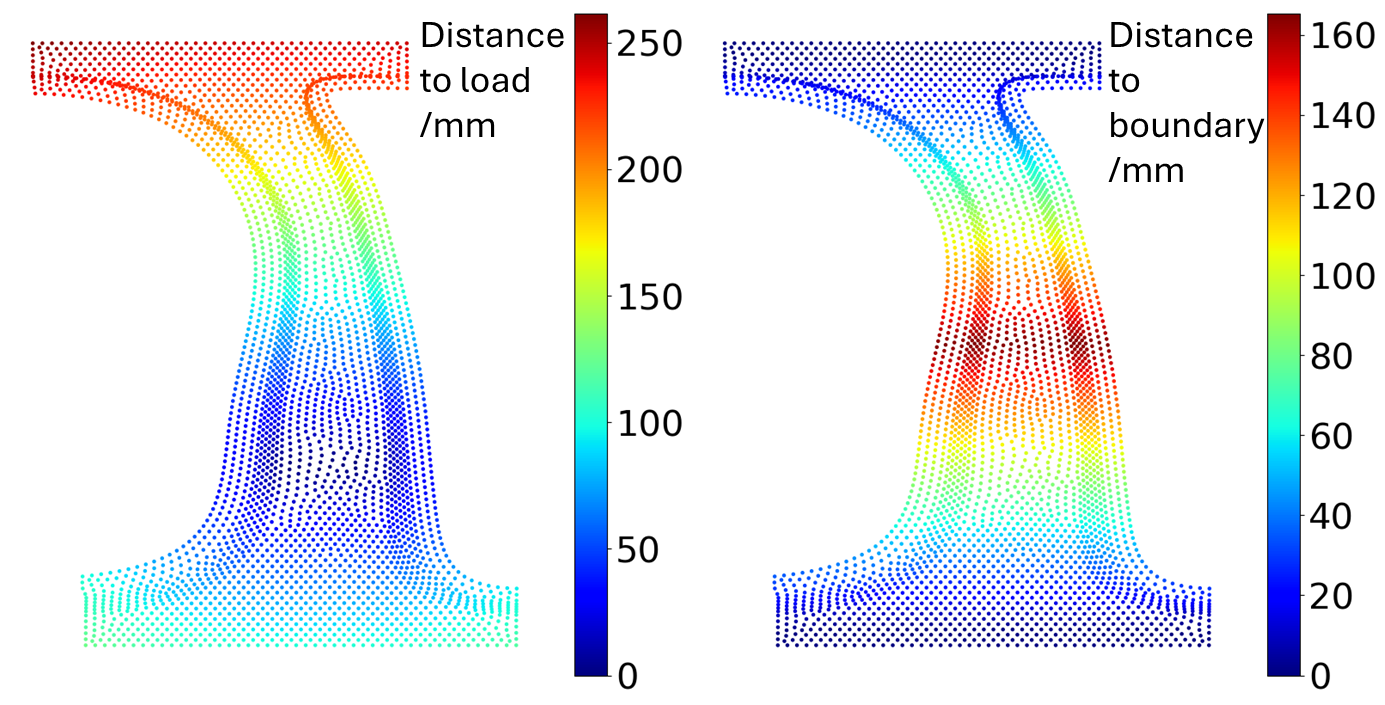}
        \\[1mm]
        {\small (c)}
    \end{minipage}
    \caption{Illustration of positional encodings: (a) the first five Laplacian eigenvectors (LE); (b) one-hot node types (1H), where the three subfigures correspond to free nodes, boundary nodes, and contact nodes, respectively; and (c) distance-to-critical-region channels (DTC), where the two subfigures correspond to distance to the loading region and distance to the boundary, respectively.}
    \label{fig:positional_encoding_examples}
\end{figure}

Figure~\ref{fig:positional_encoding_examples} illustrates representative positional encodings used in this study. Figure~\ref{fig:positional_encoding_examples}(a) shows the first five Laplacian eigenvectors (LE) of an example component, Figure~\ref{fig:positional_encoding_examples}(b) shows the one-hot node-type encoding (1H), and Figure~\ref{fig:positional_encoding_examples}(c) shows the two distance-to-critical-region channels (DTC). For DTC, the value assigned to each node is computed as its distance to the nearest node belonging to the corresponding critical region.

These encoding components are expected to provide complementary information to the model. LE provides an intrinsic description of the graph structure by representing each node in a low-frequency spectral basis of the mesh connectivity. This can help the model distinguish nodes according to their global topological location, especially when local edge features alone are insufficient to identify long-range structural context. The one-hot node-type encoding explicitly informs the model of the physical role of each node under the prescribed loading and boundary conditions, allowing it to distinguish constrained, loaded, and freely deforming regions. DTC further introduces a specific geometric descriptor by quantifying each node's proximity to the main source of deformation and to the constrained supports. This is expected to help the model learn spatially varying crash responses, such as local indentation near the impactor and the decay of deformation towards the fixed boundaries. Overall, these positional features are intended to complement the edge-based geometric representation by combining intrinsic graph position, boundary/loading semantics, and proximity to critical regions.

Table~\ref{tab:positional_encoding_ablation} summarises the results for different positional-encoding combinations. Although LE improves performance compared with Zeros and 1H, combining LE with 1H + DTC leads to worse results than using 1H + DTC alone. This may be because of several factors. First, Laplacian eigenvectors are subject to sign ambiguity, since each eigenvector can be multiplied by $-1$ while representing the same spectral mode. As a result, the same geometric region may receive inconsistent signs across different samples, making it harder for the model to learn a stable correspondence between the spectral channels and the physical deformation response. Second, when the component geometry varies across samples, the ordering and spatial pattern of higher-order eigenvectors may also change, particularly when eigenvalues are close. This can introduce sample-dependent variations in the LE channels that are not directly related to the crash response. Finally, LE increases the node-feature dimension, which introduces additional computational cost and may make optimisation more difficult under the same training budget. In contrast, the 1H + DTC setting provides more directly interpretable and problem-specific information. The node type identifies the loading and boundary regions, while the distance features describe each node's proximity to these critical regions. This configuration achieves the best performance on both metrics and is therefore used as the default node-feature configuration in the rest of this paper.

\begin{table}[H]
    \centering
    \caption{Positional-encoding ablation on B-pillar A3. Best values are shown in bold.}
    \label{tab:positional_encoding_ablation}
    \begin{tabular}{lcc}
        \toprule
        Positional encoding & MED (mm) & MIPE (\%) \\
        \midrule
        Zeros & 0.2208 & 1.1493 \\
        1H & 0.1878 & 0.7416 \\
        LE & 0.1563 & 0.5320 \\
        1H + DTC & \textbf{0.1261} & \textbf{0.3536} \\
        LE + 1H + DTC & 0.1348 & 0.4303 \\
        \bottomrule
    \end{tabular}
\end{table}

\subsubsection{Mask ratio}
\label{Mask ratio}

\begin{figure}[H]
    \centering
    \begin{minipage}[t]{0.49\linewidth}
        \centering
        \includegraphics[width=\linewidth]{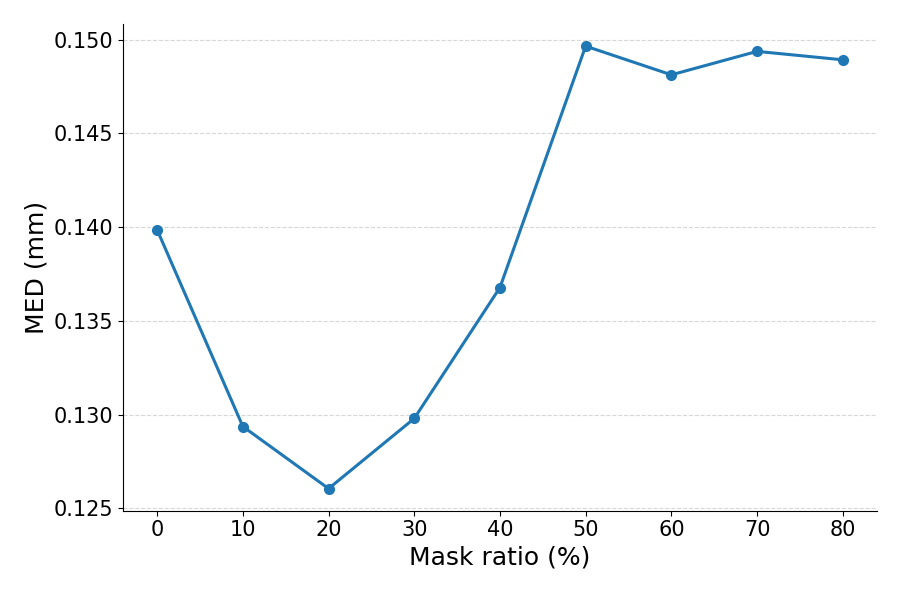}\\[0.5mm]
        {\small (a)}
    \end{minipage}\hfill
    \begin{minipage}[t]{0.49\linewidth}
        \centering
        \includegraphics[width=\linewidth]{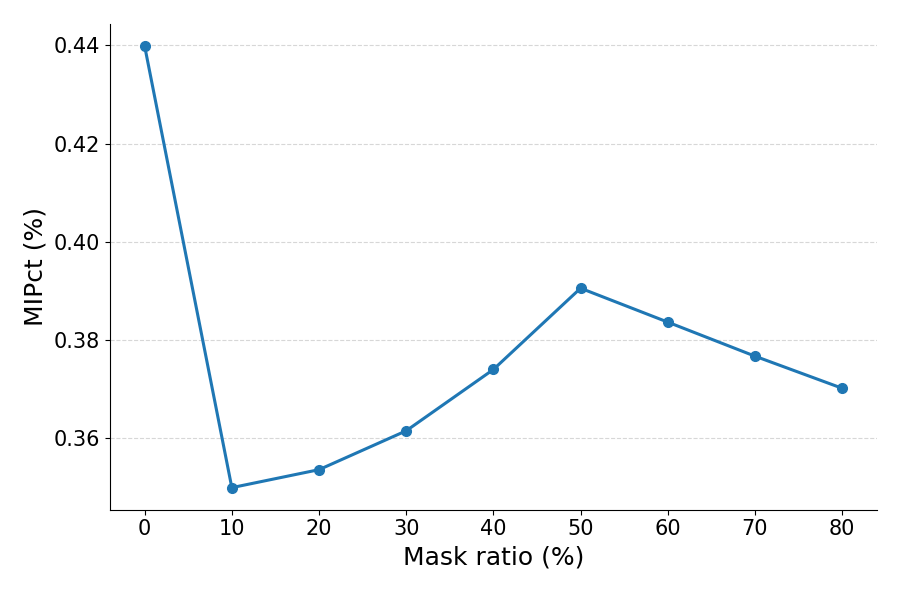}\\[0.5mm]
        {\small (b)}
    \end{minipage}
    \caption{Mask-ratio ablation: (a) MED (mm) versus mask ratio and (b) MIPE (\%) versus mask ratio.}
    \label{fig:mask_ratio_ablation}
\end{figure}

To quantify the effect of masking, we conduct a mask ratio ablation and evaluate the MGUNet model with mask ratios from 0\% to 80\%. We use B-pillar A3 case for this ablation study. Figure~\ref{fig:mask_ratio_ablation} reports the corresponding MED and MIPE.

The results show that introducing moderate masking improves generalisation compared with no masking (0\%). Performance is best in the lower masking range. MED reaches its minimum at 20\%, while MIPE is lowest at 10\% and remains competitive at 20\%. When the mask ratio further increases, both metrics degrade, indicating that excessive information removal harms displacement prediction quality. Considering the overall trade-off across both metrics, we select a 20\% mask ratio for all subsequent experiments in this paper.

\subsection{Baseline comparisons}
\label{Baseline comparisons}

We benchmark MMGUNet on two case studies, \emph{B-pillar A3} and \emph{U-channel}, trained using six different models in total. These case studies are chosen to cover two complementary levels of geometric difficulty. As summarised in Table~\ref{tab:case_study_parameters}, B-pillar A3 represents a relatively simple case with lower shape complexity but still sufficiently large shape variation. The U-channel represents a more challenging case with the largest design space variation, such that samples differ more strongly from one another. The three external baseline models are MGN \cite{Pfaff2021MeshGraphNets}, MS-MGN \cite{Fortunato2022MultiscaleMeshGraphNets}, and Multigrid \cite{Garnier2024MultiGridSelfAttention}. For MGN, the model hyperparameters follow the settings suggested in the original paper. MS-MGN is implemented as a hierarchical variant with three downsampling layers to provide a fair comparison with our multiscale setting. Multigrid is implemented as a W-cycle model with self-attention pooling, again using the parameter settings suggested in the original work. The remaining three models serve as model ablations: GUNet-fix, MGUNet, and the final MMGUNet. GUNet-fix constructs cross-graph edge connections directly from the shared coarsened graphs without morphing, such that the coarsened graphs generally remain geometrically mismatched with the input graphs. MGUNet instead establishes the cross-graph edges after morphing the coarsened graphs to match the shape of each input graph. MMGUNet further extends MGUNet by incorporating masked pretraining and parameter-efficient fine-tuning. The comparison isolates the two mechanisms needed for large-variation performance. MGUNet tests whether morphing the coarse hierarchy improves geometric correspondence while preserving fixed topology, thereby improving no-retraining generalisability. MMGUNet then tests whether masked supervised pretraining further improves robustness and transferability. 

\begin{figure}[t]
    \centering
    \begin{minipage}[t]{0.49\linewidth}
        \centering
        \includegraphics[width=\linewidth]{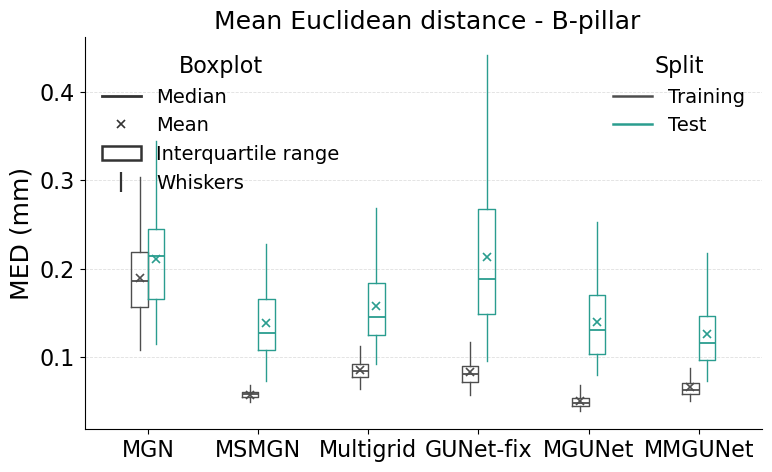}
        \par\vspace{1mm}
        {\small (a)}
    \end{minipage}
    \hfill
    \begin{minipage}[t]{0.49\linewidth}
        \centering
        \includegraphics[width=\linewidth]{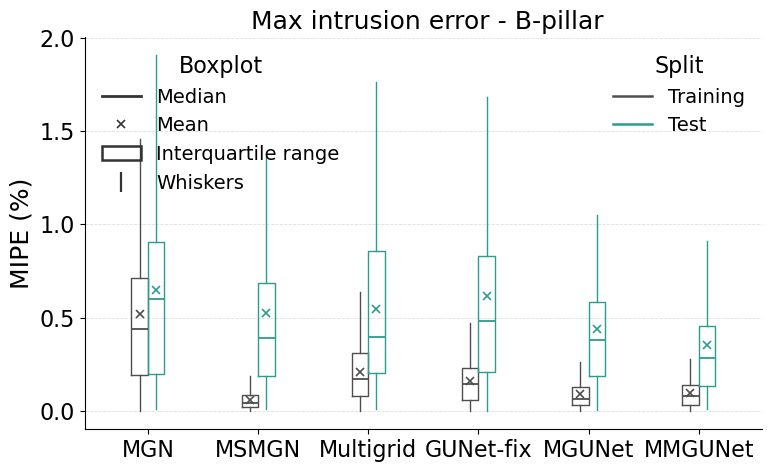}
        \par\vspace{1mm}
        {\small (b)}
    \end{minipage}
    \caption{Baseline comparison on the B-pillar A3 case study. (a) Distribution of MED (mm). (b) Distribution of MIPE (\%). For each model, grey and teal boxplots denote training and test distributions, respectively.}
    \label{fig:baseline_bpillar_xyz}
\end{figure}

Table~\ref{tab:baseline_performance_summary} shows that MMGUNet achieves the strongest overall prediction accuracy on both cases, and Figure~\ref{fig:baseline_bpillar_xyz} shows that it attains the lowest mean error on the B-pillar case with a comparatively tight error spread. A useful way to interpret these results is through the trade-off between message-propagation capacity and generalisation. MGN exhibits a relatively small train--test gap, but its absolute performance remains limited, which is likely due to insufficient message passing steps on the fine graph. By contrast, MS-MGN, Multigrid, and GUNet-fix partially relieve this bottleneck by introducing multiscale pathways, but these richer mechanisms also make the models more likely to overfit, leading to a larger train--test discrepancy. For GUNet-fix in particular, the problem is compounded by the use of fixed downsampled graphs. When shape variation is large, the resulting cross-graph edge connections become geometrically mismatched, which weakens information propagation between graph levels and leads to poorer test performance \cite{Li2025ReGUNet}. Comparing GUNet-fix and MGUNet, the benefit of coarsened-graph morphing is clear, as MGUNet achieves substantially lower errors by establishing more meaningful cross-graph edge connections. However, MGUNet still exhibits a relatively large generalisation gap, suggesting that improved cross-graph alignment alone is insufficient to fully prevent overfitting. Relative to MGUNet, MMGUNet further reduces both the error median and mean while maintaining compact interquartile ranges, indicating that the gain is not only in average accuracy but also in distributional robustness across samples. This improved behaviour is because masked pretraining regularises representation learning, while parameter-efficient fine-tuning constrains task adaptation to a small subset of parameters, thereby reducing the tendency to overfit while preserving the transferable multiscale propagation patterns learned during pretraining.

\begin{table*}[t]
    \centering
    \caption{Performance metrics summary on B-pillar and U-channel case studies (lower is better). Results are reported as mean $\pm$ standard deviation, and best values are shown in bold.}
    \label{tab:baseline_performance_summary}
    \footnotesize
    \resizebox{\textwidth}{!}{%
    \begin{tabular}{@{}lcccc@{}}
        \toprule
         & \multicolumn{2}{c}{B-pillar} & \multicolumn{2}{c}{U-channel} \\
        \cmidrule(lr){2-3} \cmidrule(lr){4-5}
        Model & MED (mm) & MIPE (\%) & MED (mm) & MIPE (\%) \\
        \midrule
        MGN \cite{Pfaff2021MeshGraphNets} & 0.2114 $\pm$ 0.0548 & 0.6494 $\pm$ 0.5428 & 2.7048 $\pm$ 0.8139 & 7.4316 $\pm$ 6.8366 \\
        MS-MGN \cite{Fortunato2022MultiscaleMeshGraphNets} & 0.1385 $\pm$ 0.0381 & 0.5243 $\pm$ 0.5087 & 2.3049 $\pm$ 0.6511 & 7.3050 $\pm$ 7.1414 \\
        Multigrid \cite{Garnier2024MultiGridSelfAttention} & 0.1573 $\pm$ 0.0473 & 0.5499 $\pm$ 0.4237 & 2.1793 $\pm$ 0.6032 & 4.9226 $\pm$ 4.4112 \\
        GUNet-fix \cite{Li2025ReGUNet} & 0.2126 $\pm$ 0.0843 & 0.6195 $\pm$ 0.5744 & 1.7956 $\pm$ 0.5484 & 3.7589 $\pm$ 3.0960 \\
        MGUNet & 0.1399 $\pm$ 0.0432 & 0.4398 $\pm$ 0.3303 & 1.5441 $\pm$ 0.5536 & 3.1748 $\pm$ 2.8152 \\
        MMGUNet & \textbf{0.1261} $\pm$ 0.0389 & \textbf{0.3536} $\pm$ 0.3037 & \textbf{1.5177} $\pm$ 0.5256 & \textbf{2.4113} $\pm$ 2.1021 \\
        \bottomrule
    \end{tabular}%
    }
\end{table*}

\begin{figure}[t]
    \centering
    \begin{minipage}[t]{0.49\linewidth}
        \centering
        \includegraphics[width=\linewidth]{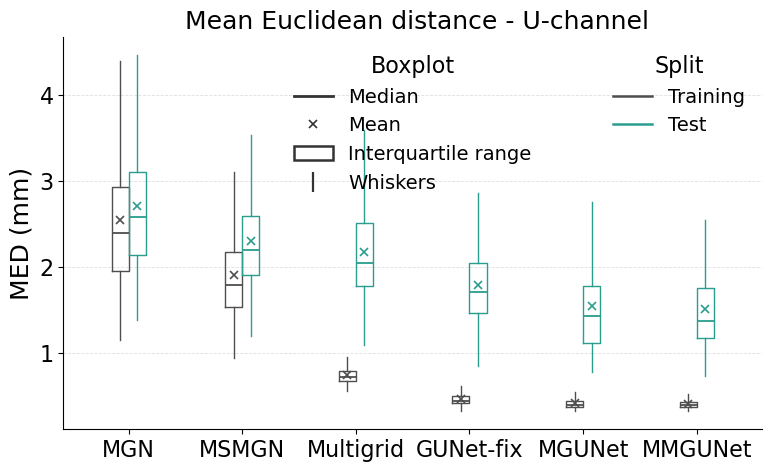}
        \par\vspace{1mm}
        {\small (a)}
    \end{minipage}
    \hfill
    \begin{minipage}[t]{0.49\linewidth}
        \centering
        \includegraphics[width=\linewidth]{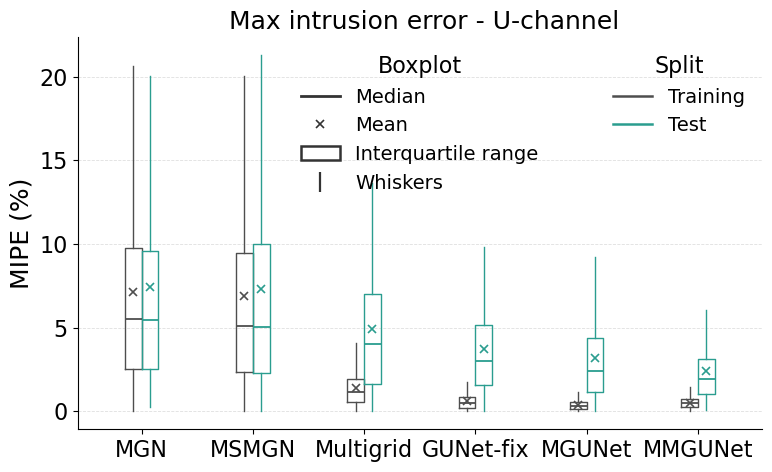}
        \par\vspace{1mm}
        {\small (b)}
    \end{minipage}
    \caption{Baseline comparison on the U-channel case study. (a) Distribution of MED (mm). (b) Distribution of MIPE (\%). Training/test splits are shown by grey/teal boxplots.}
    \label{fig:baseline_uchannel}
\end{figure}

The U-channel case constitutes a more challenging case study, exhibiting larger shape variation and a greater deformation scale than B-pillar A3. Therefore larger absolute errors for all methods can be observed in Figure~\ref{fig:baseline_uchannel}. However, similar ranking trend remains. The external baselines present the highest mean prediction errors and broadest tails, indicating limited robustness. Among the ablations, MGUNet improves generalisation over GUNet-fix. MMGUNet further shifts the test distributions downward with comparatively tighter dispersion, especially in terms of maximum intrusion prediction. Therefore, across both B-pillar A3 and U-channel case studies, MMGUNet achieves the lowest test errors among the evaluated models and shows a smaller train-test discrepancy.

\subsection{Generalisability and transferability across cases}
\label{Cross-shape}
This subsection presents the generalisation results across the four trial groups under the four training protocols defined in Section~\ref{Training strategy} using MMGUNet. We report trial-wise comparisons to evaluate how different training protocols affect cross-case generalisation performance. 

\begin{figure}[t]
    \centering
    \begin{minipage}[t]{0.99\linewidth}
        \centering
        \IfFileExists{figures/trial_training_euc.png}{%
            \includegraphics[width=\linewidth]{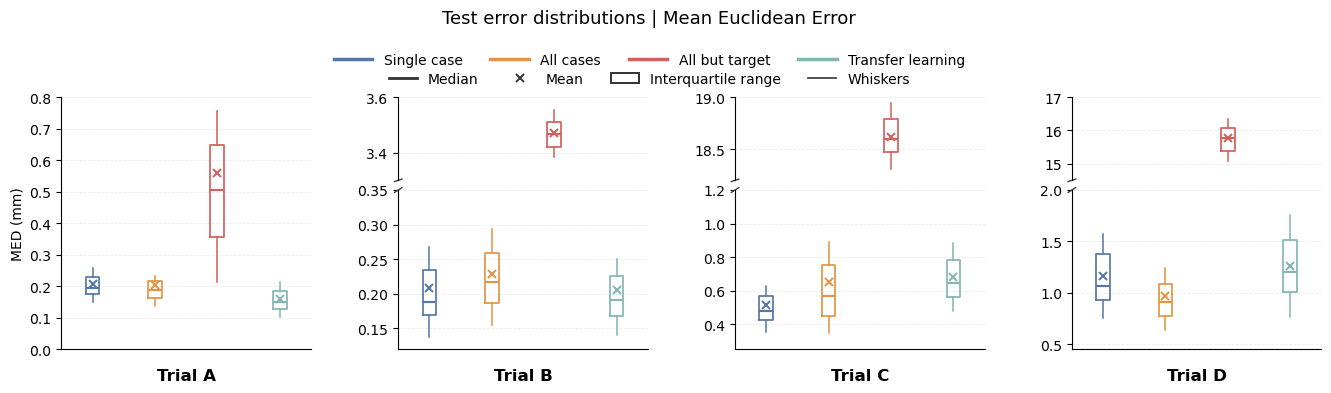}%
        }{%
            \fbox{\parbox[c][38mm][c]{0.99\linewidth}{\centering Placeholder: trial training Euclidean-distance comparison figure}}%
        }
        \\[1mm]
        {\small (a)}
    \end{minipage}\\[2mm]
    \begin{minipage}[t]{0.99\linewidth}
        \centering
        \IfFileExists{figures/trial_training_max.png}{%
            \includegraphics[width=\linewidth]{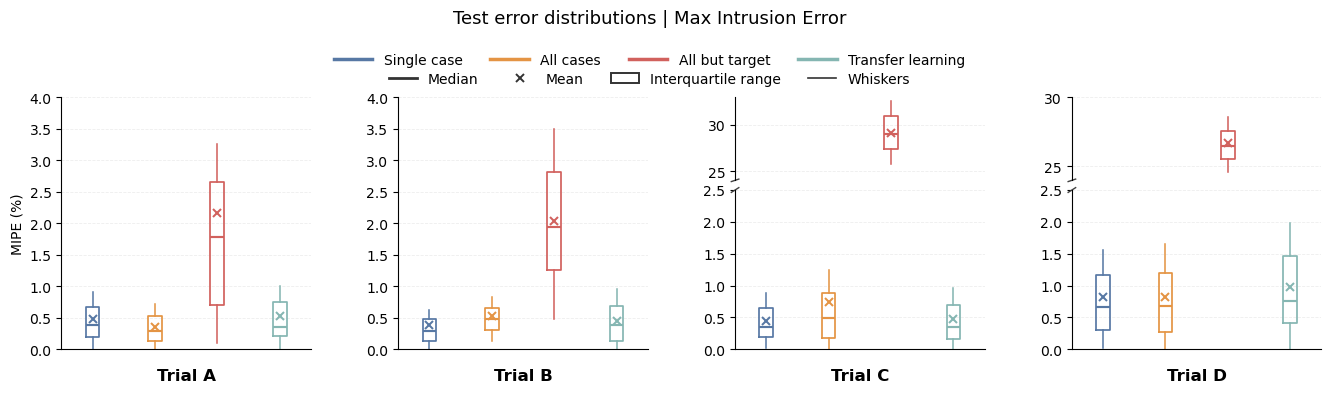}%
        }{%
            \fbox{\parbox[c][38mm][c]{0.99\linewidth}{\centering Placeholder: trial training maximum-intrusion-error comparison figure}}%
        }
        \\[1mm]
        {\small (b)}
    \end{minipage}
    \caption{Trial-wise training-strategy comparison: (a) MED and (b) MIPE. Each trial reports the four training protocols for direct comparison. Broken y-axes are used where necessary to display the substantially larger \textit{all-but-target} errors}
    \label{fig:trial_training_comparison}
\end{figure}

Figure~\ref{fig:trial_training_comparison} compares the training outcomes across all four B-pillar trials (A--D) using MED and MIPE. In each trial, we visualise the four training protocols to provide a consistent comparison.

As shown in Figure~\ref{fig:trial_training_comparison}, a consistent pattern emerges. Direct out-of-distribution generalisation (\textit{all but target}) remains challenging, especially in trials with larger distribution shifts (Trials B--D). The all-cases protocol yields performance broadly comparable to target-only training. It provides small improvements in some trials while maintaining similarly strong accuracy in others, showing that the model can be trained jointly on multiple related cases while maintaining strong predictive performance across them. For the transfer learning case, the models are fine-tuned with 300 samples to ensure fair comparisons with the other training protocols. Transfer learning consistently improves robustness and achieves better or at least comparable accuracy relative to target-only training. For simpler tasks such as Trial A, where the target case is the same component as the source cases, transfer learning achieves better prediction accuracy compared with all other training strategies. For trials with larger distribution shifts, transfer learning leads to a slightly higher but comparable prediction error compared with target-only and all-cases training. When the target problem belongs to a similar component family and shares comparable simulation conditions, boundary conditions, and material modelling assumptions, the pretrained model can be reused effectively. Under this regime, the transfer strategy is also more training-efficient, requiring fewer optimisation steps and less wall-clock training time to reach similar or better performance levels.

\begin{figure}[H]
    \centering
    \begin{minipage}[t]{0.98\linewidth}
        \centering
        \IfFileExists{figures/transfer_trialB_euc.png}{%
            \includegraphics[width=\linewidth]{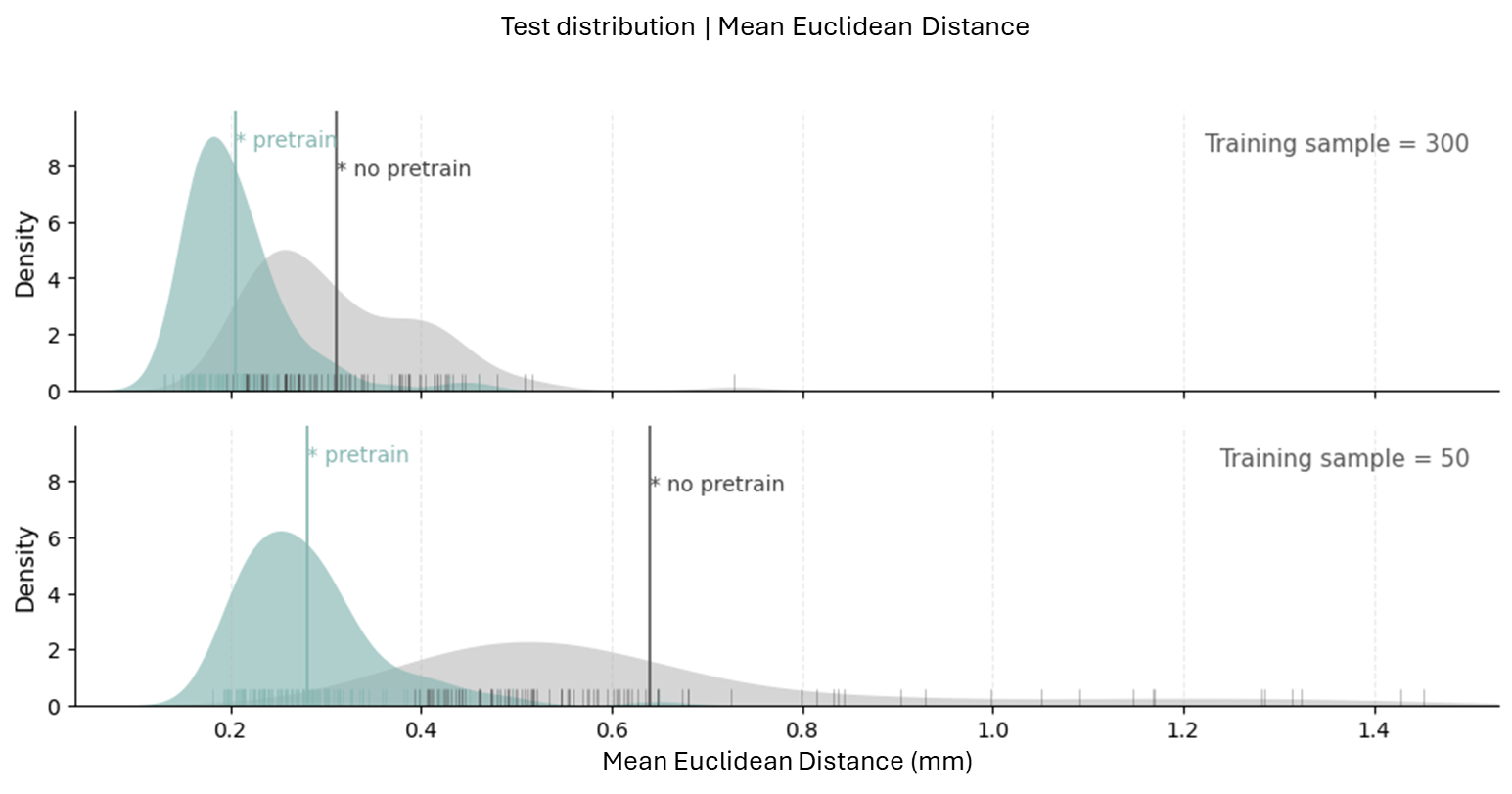}%
        }{%
            \fbox{\parbox[c][38mm][c]{0.95\linewidth}{\centering Placeholder: Trial B transfer-learning data requirement (Euclidean distance)}}%
        }
        \\[1mm]
        {\small (a)}
    \end{minipage}\\[2mm]
    \begin{minipage}[t]{0.98\linewidth}
        \centering
        \IfFileExists{figures/transfer_trialB_max.png}{%
            \includegraphics[width=\linewidth]{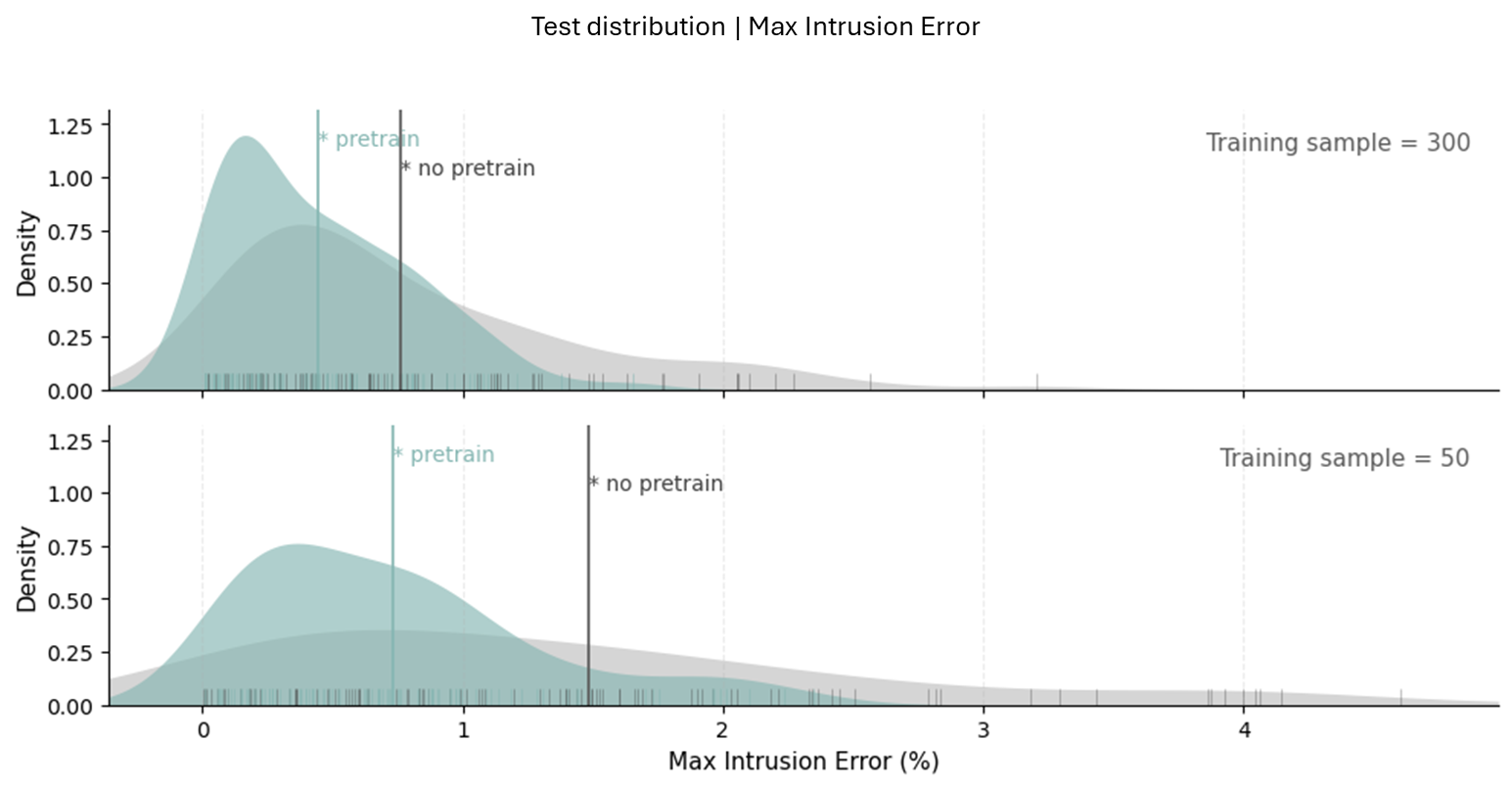}%
        }{%
            \fbox{\parbox[c][38mm][c]{0.95\linewidth}{\centering Placeholder: Trial B transfer-learning data requirement (maximum intrusion error)}}%
        }
        \\[1mm]
        {\small (b)}
    \end{minipage}
    \caption{Trial B (shape transfer learning) data-requirement analysis: (a) MED and (b) MIPE, comparing pretrained fine-tuning and no-pretrain baselines under different target-data sizes. The y-axis shows probability density, with each curve normalised to unit area, taller peaks indicate errors are more concentrated around that value.}
    \label{fig:transfer_data_requirement}
\end{figure}

\begin{figure}[H]
    \centering
    \begin{minipage}[t]{0.48\linewidth}
        \centering
        \IfFileExists{figures/vis_a.png}{%
            \includegraphics[width=\linewidth]{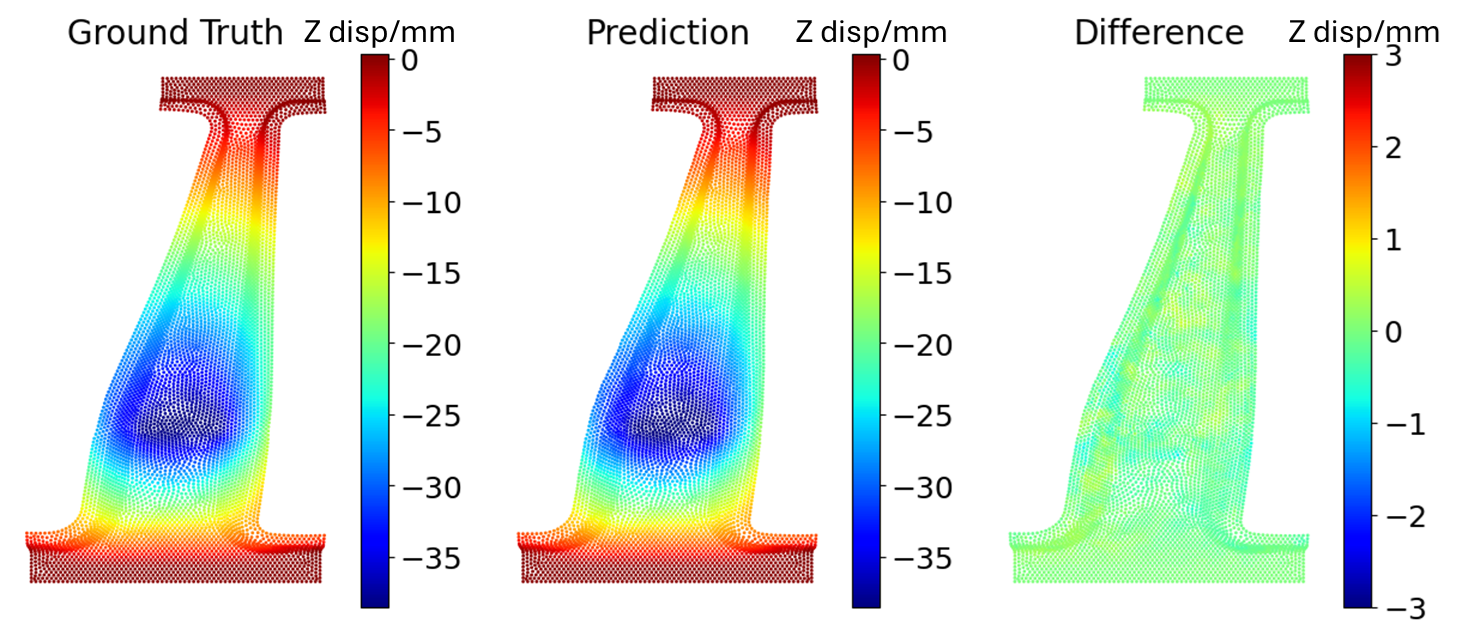}%
        }{%
            \fbox{\parbox[c][38mm][c]{0.95\linewidth}{\centering Placeholder: pretrain 300 samples}}%
        }
        \\[1mm]
        {\small (a) pretrained: fine-tune with 300 samples}
    \end{minipage}
    \hfill
    \begin{minipage}[t]{0.48\linewidth}
        \centering
        \IfFileExists{figures/vis_b.png}{%
            \includegraphics[width=\linewidth]{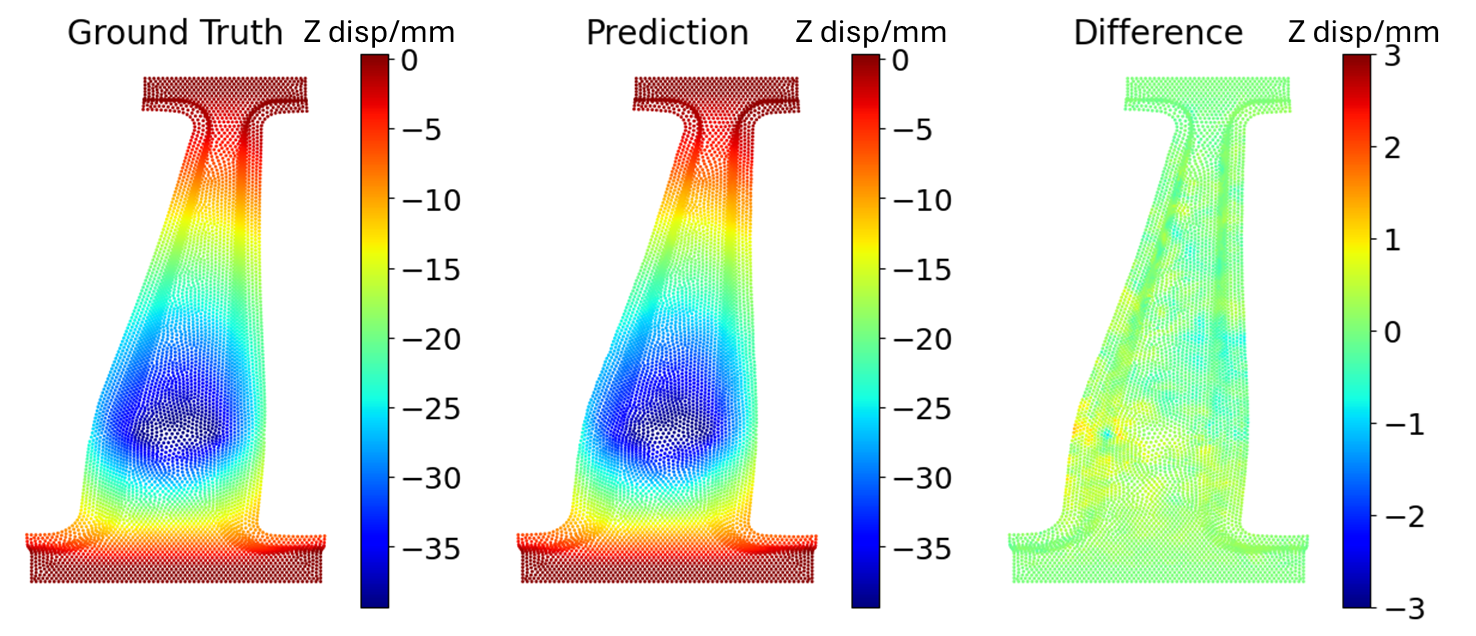}%
        }{%
            \fbox{\parbox[c][38mm][c]{0.95\linewidth}{\centering Placeholder: pretrain 50 samples}}%
        }
        \\[1mm]
        {\small (b) pretrained: fine-tune with 50 samples}
    \end{minipage}

    \vspace{2mm}

    \begin{minipage}[t]{0.48\linewidth}
        \centering
        \IfFileExists{figures/vis_c.png}{%
            \includegraphics[width=\linewidth]{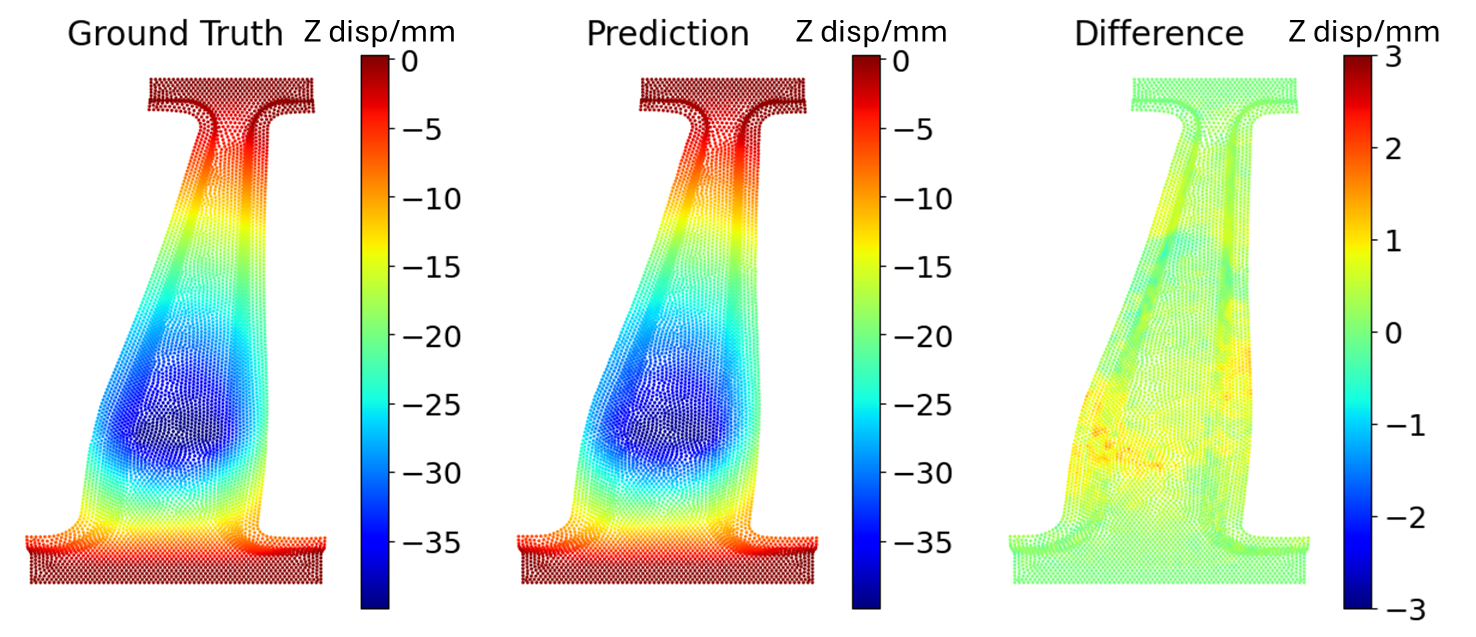}%
        }{%
            \fbox{\parbox[c][38mm][c]{0.95\linewidth}{\centering Placeholder: no pretrain 300 samples}}%
        }
        \\[1mm]
        {\small (c) train from scratch with 300 samples}
    \end{minipage}
    \hfill
    \begin{minipage}[t]{0.48\linewidth}
        \centering
        \IfFileExists{figures/vis_d.png}{%
            \includegraphics[width=\linewidth]{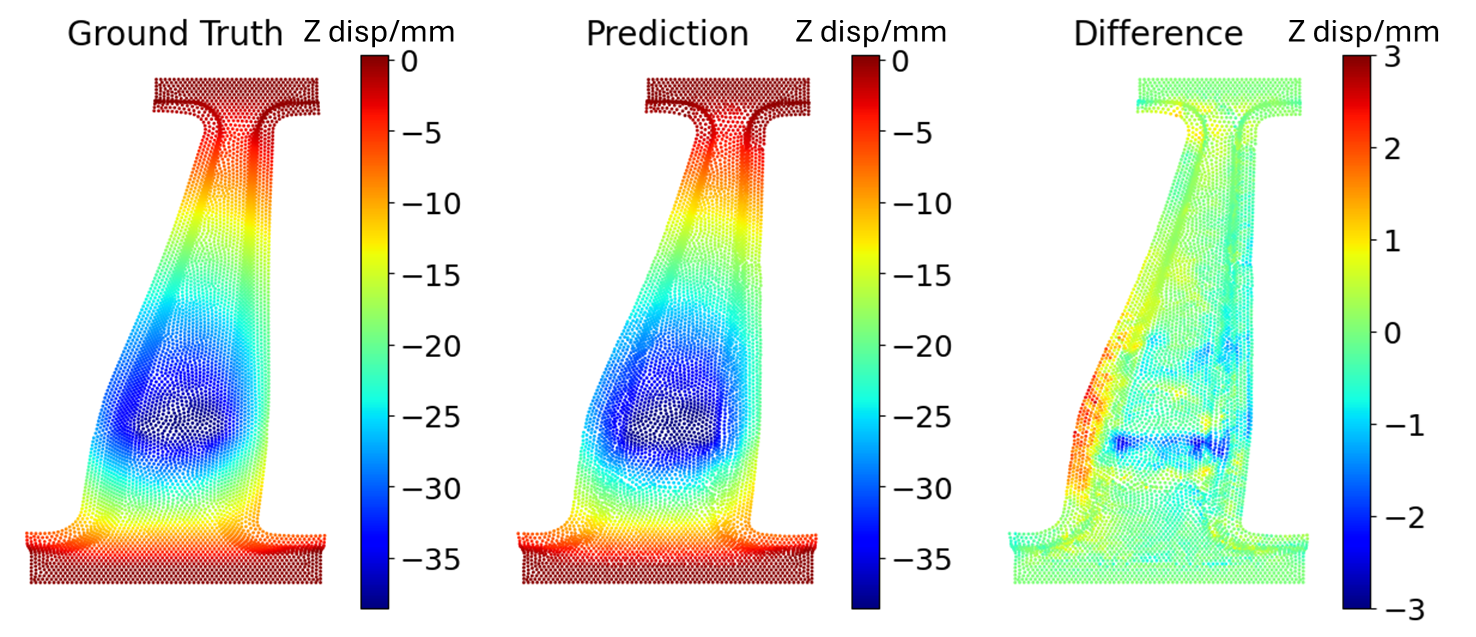}%
        }{%
            \fbox{\parbox[c][38mm][c]{0.95\linewidth}{\centering Placeholder: no pretrain 50 samples}}%
        }
        \\[1mm]
        {\small (d) train from scratch with 50 samples}
    \end{minipage}
    \caption{Trial B qualitative comparison of ground-truth and predicted $z$-displacement fields for pretrained fine-tuning and training from scratch under two target-data budgets. Each group shows ground truth, prediction, and pointwise prediction error.}
    \label{fig:pretrain_visualisation_comparison}
\end{figure}

We further analyse data efficiency for transfer learning with varying fine-tuning-set sizes. Figure~\ref{fig:transfer_data_requirement} shows an example result using Trial B. We compare the distribution of the test errors using pretrained and non-pretrained models with different numbers of training samples. As defined in Section~\ref{Generalisability study training strategy}, the pretrained transfer-learning model is first pretrained for 1000 epochs and then fine-tuned for a further 1000 epochs, whereas the no-pretrain model is trained from scratch for 1000 epochs only, without masking-based pretraining or parameter freezing. Therefore, the no-pretrain setting should not be interpreted as equivalent to the \textit{single-case} training protocol. This comparison evaluates target-data efficiency rather than equal total training compute, because the pretrained model benefits from a reusable trained backbone. The key result is that, once the backbone has been pretrained, adaptation to a new target case requires substantially fewer target samples than training from scratch. Moreover, fine-tuning with only 50 target samples can outperform no-pretrain baseline with 300 samples, demonstrating that pretraining significantly reduces target-data requirements while improving final accuracy. Figure~\ref{fig:pretrain_visualisation_comparison} further provides a qualitative comparison between ground truth and prediction under these four settings. Fine-tuning a pretrained model with 300 samples shows the closest agreement with the ground truth, while fine-tuning with only 50 samples still shows slightly better agreement than training from scratch with 300 samples. By contrast, training from scratch with 50 samples exhibits the poorest agreement and the largest visible prediction errors. This indicates that a backbone MMGUNet model pretrained with multiple datasets is reusable for new cases within a similar component family, and that fine-tuning requires far fewer training samples to reach comparable accuracy.

Table~\ref{tab:overall_summary} summarises the overall effect of pretraining across all trials by averaging the results obtained under the available training sample sizes within each trial. For Trials A--D, the reported averaged MED and MIPE values are computed as the mean of the results from the two sample sizes, 300 and 50. For Trial E, we evaluate one training sample size (200 samples). The percentage improvement of the pretrained model is computed relative to the no-pretrain baseline as the reduction in error:
\begin{equation}
\mathrm{Improvement}(\%) = \frac{\mathrm{E}_{\mathrm{no\text{-}pretrain}} - \mathrm{E}_{\mathrm{pretrain}}}{\mathrm{E}_{\mathrm{no\text{-}pretrain}}} \times 100\%,
\end{equation}
where $\mathrm{E}$ is the prediction error. A positive percentage therefore indicates that pretraining reduces the prediction error.

Overall, Table~\ref{tab:overall_summary} shows that pretraining consistently improves performance for both MED and MIPE across all trials. The largest gains are observed in Trials A and B, where pretraining reduces MED by 58.73\% and 48.86\%, respectively, and reduces MIPE by 48.11\% and 47.85\%, respectively. These results indicate that pretraining provides substantial benefits in easier transfer settings, with around half of the baseline error reduced. Trial C also shows clear improvement, although at a more moderate level, with reductions of 25.78\% in MED and 38.90\% in MIPE. In Trial D, the improvement remains positive but becomes smaller, at 16.37\% for MED and 22.17\% for MIPE, suggesting that the benefit of pretraining diminishes as the transfer task becomes more challenging. Trial E exhibits the smallest relative gains, with reductions of 8.41\% in MED and 10.06\% in MIPE, but still demonstrates that pretraining remains beneficial even in the cross-component setting. Taken together, these results show a robust and consistent advantage of pretraining across all trials, while also indicating that the magnitude of the benefit depends on the difficulty of the transfer scenario. A detailed summary of the prediction accuracy of all trials with different fine-tuning-set budgets is presented in ~\ref{appendix:trialwise_full}.

\begin{table}[H]
\centering
\caption{Overall trial-level summary of pretraining vs. no pretraining.}
\label{tab:overall_summary}
\footnotesize
\setlength{\tabcolsep}{5pt}
\renewcommand{\arraystretch}{1.15}
\resizebox{\textwidth}{!}{%
\begin{tabular}{@{}lcccccc@{}}
\hline
Trial & \multicolumn{3}{c}{MED} & \multicolumn{3}{c}{MIPE} \\
\cline{2-7}
 & Pretrain & No pretrain & Improvement & Pretrain & No pretrain & Improvement \\
\hline
Trial A & 0.1869 & 0.4527 & 58.73\% & 0.6337 & 1.2212 & 48.11\% \\
Trial B & 0.2436 & 0.4763 & 48.86\% & 0.5862 & 1.1242 & 47.85\% \\
Trial C & 0.8639 & 1.1640 & 25.78\% & 0.7794 & 1.2755 & 38.90\% \\
Trial D & 1.4855 & 1.7763 & 16.37\% & 1.1938 & 1.5340 & 22.17\% \\
Trial E & 2.6173 & 2.8575 & 8.41\% & 6.7032 & 7.4533 & 10.06\% \\
\hline
\end{tabular}%
}
\end{table}

\section{Conclusion}
\label{Conclusion}

This paper presents Mask-Morph Graph U-Net, a generalisable mesh-based surrogate for crashworthiness field prediction under large geometric variation. The method addresses a key limitation of fixed-hierarchy, edge-specific graph surrogates: fixed coarse topology is desirable for high-capacity edge-specific aggregation, but can lead to poor fine-to-coarse correspondence when component shape, scale or complexity varies substantially. MMGUNet resolves this tension by preserving fixed coarse connectivity while morphing the coarsened graph hierarchy to each input geometry. In addition, supervised masked pretraining followed by parameter-efficient fine-tuning reduces overfitting and improves target-data efficiency in transfer learning settings.

In \textit{single-case} training, the results show that the proposed model achieves higher prediction accuracy and a smaller train--test gap than existing and ablated baselines, indicating reduced overfitting and improved robustness. In transfer scenarios, masked pretraining with parameter-efficient fine-tuning consistently outperforms no-pretrain baseline at the same data budget. In Trials A and B, fine-tuning with 50 target samples surpasses no-pretrain baseline with 300 samples. Overall, pretraining improves both MED and MIPE in all cases. The largest gains in Trial A reach 58.73\% MED reduction and 48.11\% MIPE reduction. We also observe consistent positive gains for all transfer learning case studies.

In practical crashworthiness design, engineers often need rapid estimates of deformation fields and intrusion measures for many geometric variants before committing to expensive nonlinear FE simulations. The proposed surrogate supports this workflow by predicting nodal displacement fields directly on irregular FE meshes and by reducing the amount of target data required when adapting to related geometries. The method is therefore intended as a decision-support tool for early-stage design exploration rather than a replacement for final certification-level crash simulation.

The current study is limited to terminal displacement prediction under fixed loading, material, contact, and boundary conditions. The proposed morphing strategy also assumes identifiable geometric landmarks and a shared coarse graph hierarchy across the evaluated component families. Direct prediction for strongly shifted target geometries without target-domain fine-tuning remains challenging. Future work will extend the framework to time-dependent crash responses, variable loading and material conditions, broader component families, and larger vehicle-level assemblies. Future work will also integrate the surrogate into a broader design-optimisation platform for vehicle panel components.

\section*{Acknowledgements}
The authors acknowledge funding support from UKRI (UKRI221: AI-Driven Design for Forming High-Performance Vehicle Parts), as well as PhD scholarships from Imperial College London. They would also like to thank ESI Group for providing technical support with the Virtual-Performance Solution (VPS). For the purpose of open access, the authors have applied a Creative Commons Attribution (CC BY) license to any Author Accepted Manuscript version arising.

\section*{Declaration of competing interest}
The authors declare that they have no known competing financial interests or personal relationships
that could have appeared to influence the work reported in this paper.

\bibliographystyle{elsarticle-num}
\bibliography{references}

\appendix
\clearpage
\section{Dataset parameters}
\label{Dataset parameters}

In this appendix, we provide the detailed parameter definitions used to generate the B-pillar and U-channel datasets. For the B-pillar case, as illustrated in Figure~\ref{fig:case_studies}, geometric variation is introduced by morphing the top region of the component in the $x$ and $y$ directions, as well as morphing in the $z$ direction at one of the three control points. The magnitude of each morph is defined relative to the characteristic length of the component in the corresponding direction. For example, the variation in the $x$ direction is specified as $\pm 5\%$ of the total B-pillar width, i.e. its length in the $x$ direction. The full set of B-pillar parameters and bounds is listed in Table~\ref{tab:case_study_parameters}.

\begin{figure}[!htbp]
\centering
\includegraphics[width=0.75\textwidth]{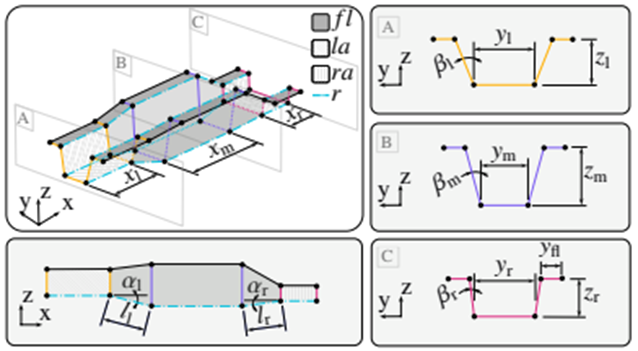}
\caption{Illustration of the parameters for U-channel 2 in the UCSM dataset \cite{Lehrer2025UCSM}.}
\label{fig:ucsm_uchannel2_parameters}
\end{figure}

\begin{figure}[!htbp]
\centering
\begin{minipage}[t]{0.64\textwidth}
    \centering
    \includegraphics[width=\linewidth]{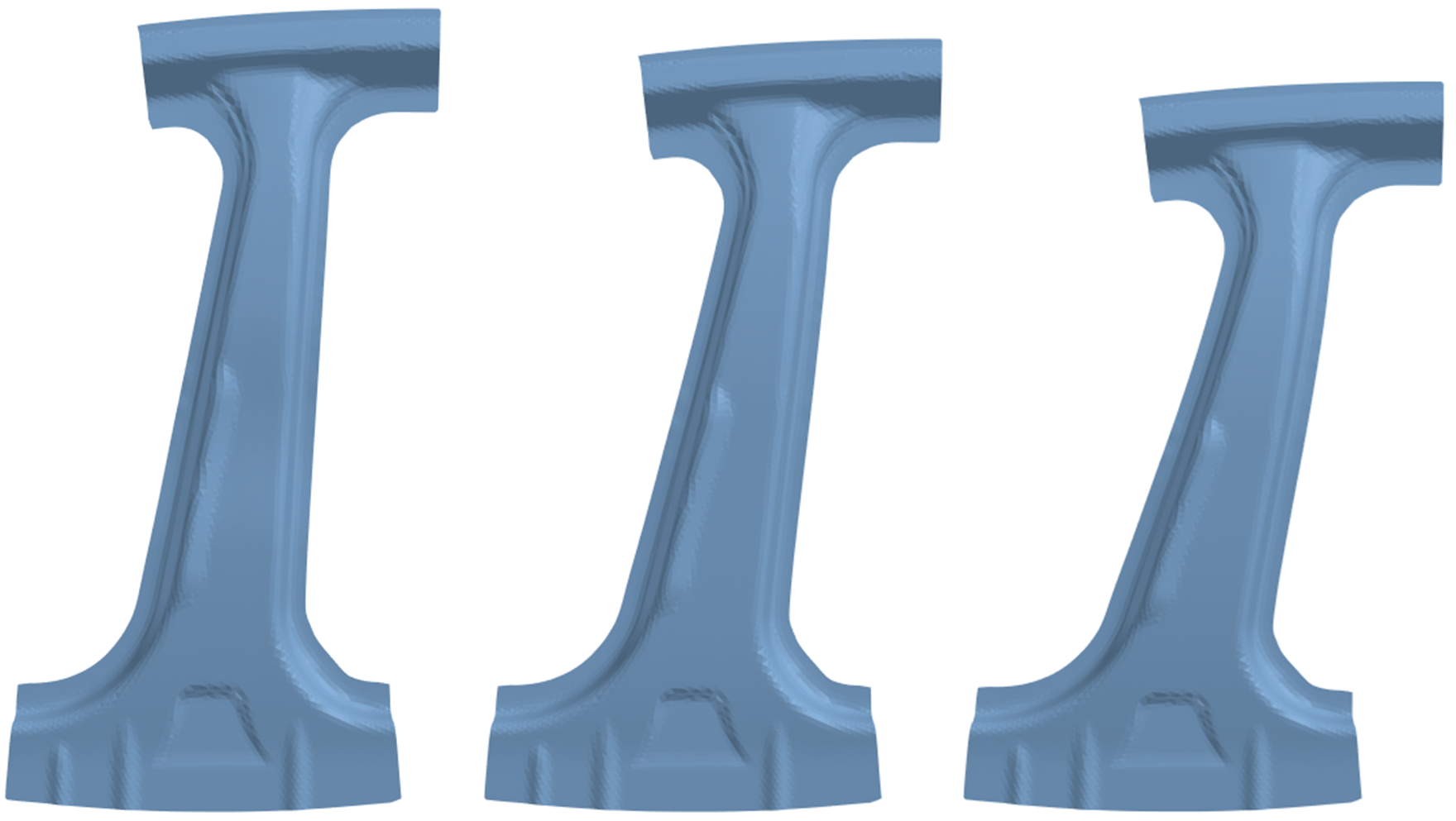}

    (a)
\end{minipage}
\hfill
\begin{minipage}[t]{0.28\textwidth}
    \centering
    \includegraphics[width=\linewidth]{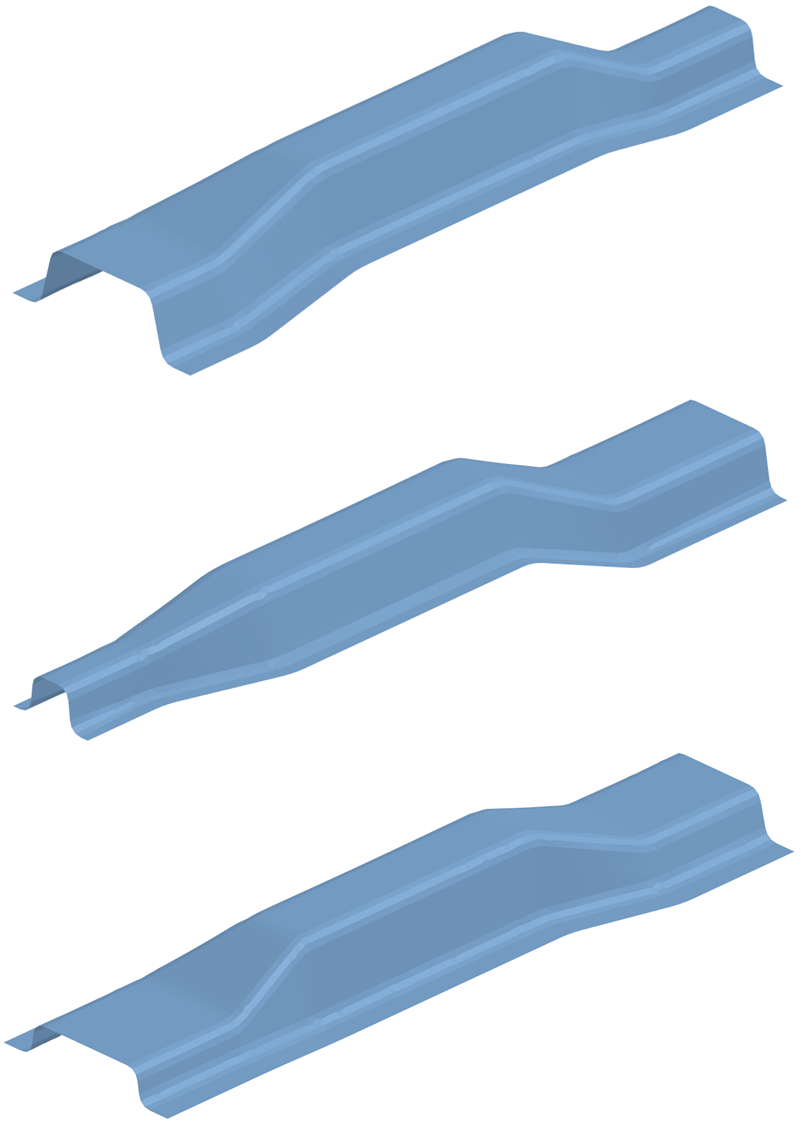}

    (b)
\end{minipage}
\caption{Representative variation of the component shapes. Figure (a) shows B-pillar variations, and Figure (b) shows U-channel variations.}
\label{fig:appendix_shape_variations}
\end{figure}

For the U-channel case, the geometry can be decomposed into three main regions: the middle plane, the left addendum, and the right addendum, as shown in Figure~\ref{fig:ucsm_uchannel2_parameters}. We set the length of the middle plane, $x_m$, to 200~mm as the reference parameter. A total of 16 geometric parameters are varied within predefined bounds, as summarised in Table~\ref{tab:case_study_parameters}. These parameters describe the lengths, widths, and heights of the different regions, as well as the slant and plane angles. Because of the larger number of parameters and their broader ranges, the U-channel case spans a wider design space than the B-pillar case and is therefore inherently more challenging for the surrogate model to learn. Representative examples of geometric variation for the B-pillar and U-channel components are shown in Figure~\ref{fig:appendix_shape_variations}.

\begin{table}[!htbp]
\centering
\caption{Case study parameters, bounds, and descriptions.}
\label{tab:case_study_parameters}
\small
\setlength{\tabcolsep}{4pt}
\begin{tabular}{p{0.16\textwidth} p{0.13\textwidth} p{0.24\textwidth} p{0.37\textwidth}}
\hline
\textbf{Case study} & \textbf{Parameter} & \textbf{Bounds} & \textbf{Description} \\
\hline
B-pillar & X morph & $x_b\cdot[-0.05,0.05]$ & Morphing top region in the $x$ direction relative to the $x$ length \\
 & Y morph & $y_b\cdot[0,0.1]$ & Morphing top region in the $y$ direction relative to the $y$ length \\
 & Z morph & $z_b\cdot[-0.1,0.1]$ & Morphing at one control point in the $z$ direction \\
 & Z-morph control point & $\{1,2,3\}$ & $Z$ morphing control point selection \\
\hline
U-channel & $x_m$ & $200$ mm & Length of the middle plane (reference parameter) \\
 & $x_l$ & $x_m\cdot[0.25,0.75]$ & Length of the left addendum \\
 & $x_r$ & $x_m\cdot[0.25,0.75]$ & Length of the right addendum \\
 & $l_l$ & $x_m\cdot[0.375,0.5]$ & Length of the left slant along its tangent \\
 & $l_r$ & $x_m\cdot[0.375,0.5]$ & Length of the right slant along its tangent \\
 & $y_m$ & $x_m\cdot[0.2,0.375]$ & Width of the middle plane \\
 & $y_l$ & $x_m\cdot[0.125,0.5625]$ & Width of the left addendum \\
 & $y_r$ & $x_m\cdot[0.125,0.5625]$ & Width of the right addendum \\
 & $y_{fl}$ & $x_m\cdot[0.05,0.15]$ & Width of the flange \\
 & $z_m$ & $x_m\cdot[0.167,0.3]$ & Height of the middle plane \\
 & $z_l$ & $x_m\cdot[0.111,0.333]$ & Height of the left addendum \\
 & $z_r$ & $x_m\cdot[0.111,0.333]$ & Height of the right addendum \\
 & $\alpha_l$ & $[0,30]^\circ$ & Angle of the left slant \\
 & $\alpha_r$ & $[0,30]^\circ$ & Angle of the right slant \\
 & $\beta_l$ & $[5,15]^\circ$ & Angle of the slant of the left plane \\
 & $\beta_r$ & $[5,15]^\circ$ & Angle of the slant of the right plane \\
 & $\beta_m$ & $[5,15]^\circ$ & Angle of the slant of the middle plane \\
\hline
\end{tabular}
\end{table}

\clearpage
\section{Barycentric parameterisation foundations}
\label{appendix:morphing_algorithm}

\paragraph{Graph setting and 3-connected planarity}
Let $G=(V,E)$ be a simple undirected graph, where vertices are mesh nodes and edges follow mesh adjacency. A graph is \emph{planar} if it admits a drawing in $\mathbb{R}^2$ with non-crossing edges. A graph is \emph{3-vertex-connected} if removing any set of at most two vertices leaves the graph connected. These concepts are standard in Tutte's spring-embedding theorem, which provides key existence, uniqueness, and non-self-intersection guarantees for barycentric embeddings with convex boundary constraints. In practice, the shell meshes used in this study are processed to satisfy the required boundary-loop ordering before parameterisation.

\paragraph{Boundary and interior vertices}
Assume the shell mesh has disk topology with a single outer boundary loop. Let
\begin{equation}
B=(b_0,\dots,b_{k-1})\subset V
\end{equation}
denote the ordered boundary cycle, and let
\begin{equation}
I:=V\setminus B
\end{equation}
denote the interior vertices. The parameterisation is obtained by prescribing UV positions on $B$ and solving for UV positions on $I$.

\paragraph{Embedding map with fixed convex boundary}
We seek an embedding (parameterisation) map
\begin{equation}
\Phi:V\rightarrow\mathbb{R}^2,\qquad i\mapsto \mathbf{u}_i=(u_i,v_i),
\end{equation}
such that boundary vertices in $B$ are mapped to a convex polygonal curve, while interior vertices satisfy a barycentric equilibrium condition. This is the classical Tutte (barycentric) embedding and can be interpreted as the equilibrium of a spring system with pinned boundary.

\paragraph{Harmonic (barycentric) condition}
For an interior vertex $i\in I$, let $N(i)$ be its one-ring neighbour set and $d_i=|N(i)|$ its degree. Using uniform barycentric weights,
\begin{equation}
w_{ij}=\begin{cases}
\dfrac{1}{d_i}, & j\in N(i),\\[4pt]
0, & \text{otherwise},
\end{cases}
\qquad\Rightarrow\qquad
\sum_{j\in V} w_{ij}=1,\ \ w_{ij}\ge 0.
\end{equation}
The discrete harmonic condition is
\begin{equation}
\mathbf{u}_i = \sum_{j\in N(i)} w_{ij}\,\mathbf{u}_j,
\qquad \forall i\in I,
\end{equation}
so each interior UV position is the weighted average of its neighbours.

\paragraph{Laplacian form and partitioned linear system}
Let $A\in\mathbb{R}^{N\times N}$ be the symmetric adjacency matrix of $G$, and define the degree matrix $D=\mathrm{diag}(d_1,\dots,d_N)$. The uniform graph Laplacian is
\begin{equation}
L:=D-A.
\end{equation}
Stack UV coordinates into $U\in\mathbb{R}^{N\times 2}$ (columns correspond to $u$ and $v$). The interior barycentric constraints become
\begin{equation}
(LU)_i=\mathbf{0}\in\mathbb{R}^2,
\qquad \forall i\in I.
\end{equation}

Impose Dirichlet boundary values $U_B\in\mathbb{R}^{|B|\times 2}$ on $B$, and denote unknown interior UVs by $U_I\in\mathbb{R}^{|I|\times 2}$. Partition indices as $(I,B)$:
\begin{equation}
U=\begin{bmatrix}U_I\\ U_B\end{bmatrix},
\qquad
L=\begin{bmatrix}L_{II} & L_{IB}\\ L_{BI} & L_{BB}\end{bmatrix}.
\end{equation}
Then the interior harmonic constraints yield the sparse linear system
\begin{equation}
L_{II}U_I + L_{IB}U_B = 0
\quad\Longrightarrow\quad
L_{II}U_I = -L_{IB}U_B,
\end{equation}
which can be solved for $u$ and $v$ coordinates.

\clearpage
\section{Full trial-wise comparison details}
\label{appendix:trialwise_full}

Tables~\ref{tab:trialwise_med} and~\ref{tab:trialwise_mipe} report the detailed trial-wise comparison between pretrained and non-pretrained models under different training sample sizes. Results are presented as mean $\pm$ standard deviation for MED and MIPE, respectively. The results show consistent accuracy improvements for the pretrained model compared with the model without pretraining.

\begin{table}[!htbp]
\centering
\caption{Full trial-wise comparison of pretrained and non-pretrained models using mean Euclidean distance (MED, mm). Results are reported as mean $\pm$ standard deviation.}
\label{tab:trialwise_med}
\normalsize
\setlength{\tabcolsep}{8pt}
\renewcommand{\arraystretch}{1.12}
\begin{tabular}{llcc}
\hline
Trial & Samples & Pretrain & No pretrain \\
\hline
Trial A & 300 & 0.1595 $\pm$ 0.0402 & 0.3182 $\pm$ 0.0856 \\
Trial A & 50  & 0.2142 $\pm$ 0.0563 & 0.5872 $\pm$ 0.2281 \\
\hline
Trial B & 300 & 0.2057 $\pm$ 0.0552 & 0.3120 $\pm$ 0.0892 \\
Trial B & 50  & 0.2815 $\pm$ 0.0748 & 0.6406 $\pm$ 0.2768 \\
\hline
Trial C & 300 & 0.6817 $\pm$ 0.1641 & 0.8188 $\pm$ 0.2509 \\
Trial C & 50  & 1.0461 $\pm$ 0.2992 & 1.5091 $\pm$ 0.3505 \\
\hline
Trial D & 300 & 1.2640 $\pm$ 0.3179 & 1.3767 $\pm$ 0.3408 \\
Trial D & 50  & 1.7070 $\pm$ 0.4421 & 2.1759 $\pm$ 0.4114 \\
\hline
Trial E & 200 & 2.6173 $\pm$ 0.8582 & 2.8575 $\pm$ 0.7583 \\
\hline
\end{tabular}
\end{table}

\begin{table}[!htbp]
\centering
\caption{Full trial-wise comparison of pretrained and non-pretrained models using maximum intrusion percentage error (MIPE, \%). Results are reported as mean $\pm$ standard deviation.}
\label{tab:trialwise_mipe}
\normalsize
\setlength{\tabcolsep}{8pt}
\renewcommand{\arraystretch}{1.12}
\begin{tabular}{llcc}
\hline
Trial & Train & Pretrain & No pretrain \\
\hline
Trial A & 300 & 0.5279 $\pm$ 0.4771 & 0.7408 $\pm$ 0.5980 \\
Trial A & 50  & 0.7394 $\pm$ 0.5763 & 1.7015 $\pm$ 1.2891 \\
\hline
Trial B & 300 & 0.4416 $\pm$ 0.3570 & 0.7625 $\pm$ 0.6435 \\
Trial B & 50  & 0.7308 $\pm$ 0.5563 & 1.4858 $\pm$ 1.3212 \\
\hline
Trial C & 300 & 0.4780 $\pm$ 0.4187 & 0.8543 $\pm$ 0.7558 \\
Trial C & 50  & 1.0807 $\pm$ 1.0011 & 1.6966 $\pm$ 1.0631 \\
\hline
Trial D & 300 & 0.9740 $\pm$ 0.7004 & 1.1700 $\pm$ 0.8302 \\
Trial D & 50  & 1.4137 $\pm$ 0.9392 & 1.8980 $\pm$ 1.1028 \\
\hline
Trial E & 200 & 6.7032 $\pm$ 5.4959 & 7.4533 $\pm$ 6.1964 \\
\hline
\end{tabular}
\end{table}

\end{document}